\documentclass[a4paper,fleqn]{cas-dc}

\usepackage[numbers]{natbib}
\def\tsc#1{\csdef{#1}{\textsc{\lowercase{#1}}\xspace}}
\tsc{WGM}
\tsc{QE}
\tsc{EP}
\tsc{PMS}
\tsc{BEC}
\tsc{DE}
\usepackage{fvextra}
\usepackage{tcolorbox}
\tcbuselibrary{breakable,skins}

\begin{document}

\shorttitle{Language Mutations Sustain Conspiracy Persistence}
\shortauthors{Cheng, Quelle, and Hale}

\title[mode=title]{Language Mutations Sustain the Persistences of Conspiracy Theories on Social Media}

\author[1]{Calvin Yixiang Cheng}[orcid=0000-0003-4897-3293]
\ead{calvin.cheng@oii.ox.ac.uk}
\corref{cor1}
\credit{Conceptualization, literature review, study design, computation, writing.}

\author[2]{Dorian Quelle}
\credit{Methodology, formal analysis, writing review and editing.}

\author[1,3]{Scott A. Hale}
% \ead{scott.hale@oii.ox.ac.uk}
\credit{Supervision, methodology, writing review and editing.}

\affiliation[1]{organization={Oxford Internet Institute, University of Oxford},
            addressline={1 St Giles'},
            city={Oxford},
            postcode={OX1 3JS},
            country={United Kingdom}}

\affiliation[2]{organization={Department of Mathematics, University of Zurich},
            city={Zurich},
            postcode={8057},
            country={Switzerland}}

\affiliation[3]{organization={Meedan},
            city = {San Francisco}, 
            postcode = {94105},
            country = {USA}}

\cortext[cor1]{Corresponding author}
\begin{abstract}
This study investigates how language mutations affect the persistent diffusion of conspiracy theories on social media. Drawing on a three-year dataset of conspiracy-related posts from X, and applying computational linguistic analysis alongside survival modelling, we find that conspiracy claims with greater semantic mutations have substantially longer lifespans. Mutations in psycholinguistic properties, including pronouns, social reference words, cognitive process terms, risk- and health- related vocabularies, are associated with extended lifespans. Mutations in actor, action and target (AAT) categories are associated with longer lifespans as well. Qualitative analysis identifies two predominant mutation patterns: simplification and assimilation, at both linguistic and AAT structural levels. Taken together, the results advance our understanding of how language mutations contribute to conspiracy persistence online and shed lights on longitudinal content moderation strategies. We argue that content moderation should consider the mutability of conspiracy claims and focus on the core claims that can address their potential variations.
\end{abstract}


\maketitle

\section{Introduction}

% The persistence of conspiracy theories on social media has been a significant societal concern. 
Conspiracy theories, defined as explanatory content that attribute social events to secret plots by malicious people \cite{douglasUnderstanding2019,vanprooijenBelief2018,uscinskiConspiracy2018}, have demonstrated surprising persistence on social media where many of them continue to re-emerge despite repeated debunking \cite{arcosResponses2022, bodnerCOVID192020,dowCOVID2021}. A notable pattern underlying this persistence is linguistic mutation during transmission: many conspiracy theories reappear in modified forms, undergoing explicit semantic and structural changes as they circulate online \cite{shahsavariConspiracy2020,shinDiffusionMisinformationSocial2018}. Prior research on persistence has primarily focused on individual psychological characteristics \cite{douglasPsychology2017, swamiConspiracist2011} or the viral diffusion of static content \cite{vosoughispread2018}. It remains unclear to what extent language mutations contribute to the persistent diffusion of conspiracy theories.\footnote{All scripts used in the analyses presented in this article is made available on the Open Science Framework (OSF) \url{https://osf.io/usqxn/?view_only=64270fbc195e475381641a146316b59e} repository to facilitate sharing and replication (Anonymous link for  peer review).}

Language mutations refer to linguistic changes in information during transmission. They occur naturally in human communication systems. As information spreads, human transmitters often introduce various language changes, such as simplifying the content, selectively emphasizing linguistic elements, or distorting semantics to align with their prior beliefs \cite{allportAnalysis1946}. \citet{campbellSystematic1958} theorized such a phenomenon of language mutations as ``systematic errors'' in communication systems, arising from human cognitive constraints in perception, memory and information processing. 

Importantly, such mutations may sustain the persistent diffusion of information. By introducing variation and novelty, mutated misleading information can reduce information fatigue and renew audience attention \cite{sunNo2023}. This issue is particularly concerning in today’s digital environments, where social media platforms enable rapid, large-scale diffusion while simultaneously offering mechanisms for modification through, for example, quoting, and user-generated variants. These affordances create fertile ground for conspiracies to proliferate, adapt, and exert long-term impact \cite{cinelliConspiracy2022, douglasUnderstanding2019}. 

Despite good theory, there is little empirical evidence quantifying the impact of language mutations on the persistent diffusion of conspiracy theories, and it is unclear what language features can predict such persistence. This paper addresses these gaps by empirically examining the relationship between language mutations in conspiracy claims and their persistence online. Using a dataset of $446,829$ COVID-19 related conspiracy posts collected over three years from X (formerly Twitter), we analysed how mutations within conspiracy claims relate to their lifespan. In this work, we focused on the persistence of claims --- a cluster of semantically similar conspiracy posts that can be debunked by a single fact-checking claim \cite{kazemiClaim2021a}. 

Mutations are examined at three analytical scales across various language granularities.  Inspired by \citet{tangherliniGenerative2018}'s generative conspiracy framework, we operationalize mutation along three structured dimensions: semantic, psycholinguistic, and actor–action–target (AAT) mutations. Semantic mutations captures drift of conspiracy claims in embedding space, providing a fine-grained quantification of language mutation. Psycholinguistic mutations refer to changes in the distribution of vocabulary reflecting underlying psychological and cognitive states, such as emotional tone, moral judgment, and power dynamics. AAT mutations measure transformations in the narrative structure of claims by identifying changes in actors, actions, and targets categories. The persistence is modelled by survival analysis. 

Our findings show that language mutations are significantly associated with greater persistence. Claims exhibiting larger semantic mutations demonstrate substantially longer lifespans, with 27\% higher likelihood to persist compared to non-mutating claims. Also, specific psycholinguistic and AAT category changes emerge as strong predictors of persistence. Moreover, our qualitative analyses reveal two prominent mutation patterns. These results advanced our understanding of how conspiracy claims adapt and persist on social media. They extend the information mutation framework to contemporary online communication contexts and offer practical implications for the detection and mitigation of persistent conspiracy theories.

\section{Language Mutations and Persistent Diffusion}

Language mutations are widely observed among people's engagements with conspiracy claims \cite[e.g.,][]{kearneyTwitter2020, shahsavariConspiracy2020, samoryGovernment2018}. These engagements can be broadly categorized into three types: duplicatory, translatory, and reductive \cite{campbellSystematic1958}.

Duplicatory transmission refers to the process where individuals intend to duplicate certain conspiracy claims. It is common during social crises such as the COVID-19 pandemic, when there were a lot of unknowns and uncertainties. For example, conspiracy claims such as ``COVID is bioweapon from Wuhan lab'' provided simple though false explanation for the situation and attracted much duplicatory transmission. Individuals may post, repost, like, or save these claims, leading to persistent diffusion. 

Translatory transmission refers to changing the modality of the input during output, but without an intended loss of complexity or information \cite{campbellSystematic1958}. Many conspiracy claims circulate in multimedia forms beyond text, including memes, pictures, and videos. Translatory transmission can occur when individuals summarise a video in a text post, create an infographic from an article, or live-tweet a real-world event. Although the modality changes, the intent is usually to preserve the core meaning of the original content. Nevertheless, this process would also be shaped by users’ pre-existing knowledge and beliefs. 

Reductive transmission represents the transformation of complex input signals into simpler outputs. It involves user interpretation and assimilation and is often manifested through behaviours such as commenting, quoting, tagging, or adding hashtags, each of which can modify or reinterpret the original content. These actions can be shaped by a range of cognitive and emotional mechanisms, including the familiarity effect (where individuals are more likely to engage with familiar-sounding content) \cite{schwarzMetacognitive2004, schwarzMetacognitive2007}, emotional contagion (where affective content elicits stronger reactions) \cite{hatfieldEmotional1993, kramerExperimental2014}, moral arousal that prompts engagement \cite{bradyEmotion2017}, and susceptibility to logical fallacies that reinforce conspiratorial thinking \cite{swamiConspiracist2011}.

These language mutations may contribute to the persistent diffusion of conspiracy claims through two primary mechanisms: introducing novelty and reviving dormant memories. First, mutations can introduce novelty into existing claims, a factor shown to be critical for faster, deeper, and wider information diffusion in digital environments \cite{vosoughispread2018}. As online stories evolve through incorporating new elements or reframing themselves \cite{imemerging2010}, such variants may enhance their survival by increasing novelty and sustaining user interest while reduce the information fatigue \cite{sunNo2023}. Second, mutations can revitalise dormant information and boost visibility. For example, outdated political rumours often re-emerge during election seasons or major sociopolitical events, repackaged with timely details to regain traction \cite{shinDiffusionMisinformationSocial2018}.

However, few studies have empirically examined the relationship between the language mutation in claims and their persistence. To address this gap, this paper aims to examine the extant to which language mutations affect the persistence of conspiracy claims online. 

\section{Quantifying Language Mutations}

We examine language mutations in conspiracy claims from three analytical perspectives: semantic similarity, psycholinguistic properties, and actor–action–target (AAT) categories. These dimensions allow us to probe mutation in different ways. Semantic similarity captures language mutations by measuring semantic drift in a vector space, operationalized as changes among sentence embeddings. Therefore, we ask the first research question: \textit{to what extent does semantic drift affect the persistence of conspiracy claims online? (RQ1)}

Next, we investigate the changes in psycholinguistic properties of conspiracy claims, which captures the word-level granularity. As one of the most widely used approaches for analysing everyday language use \cite{chungWhat2018}, psycholinguistic properties refer to vocabularies that reflect underlying psychological and cognitive processes \cite{rainsPsycholinguistic2023}. Prior studies have shown that conspiracy claims exhibit distinct psycholinguistic patterns compared to non-conspiracy claims on social media \cite[e.g.,][]{kleinPathways2019, mianiLOCO2022, rainsPsycholinguistic2023}, offering valuable insight into how language mutations may influence their persistence.

For instance, conspiracy claims display a pronounced lexical emphasis on power-related vocabulary. Power asymmetries and conflicts, which are among the central themes in conspiracy theories, are frequently expressed through references to institutional actors (e.g., government, law), hierarchical structures (e.g., dominance, control), and extreme actions (e.g., crime, kill, hate) \cite{douglasWhat2023, kleinPathways2019,lewandowskyConspiracy2020}. Conspiracy claims also tend to contain a higher proportion of emotionally charged language, particularly words associated with negative affect. Terms expressing anger, profanity, and other negative emotions are significantly more prevalent in conspiratorial than non-conspiratorial content \cite{kleinPathways2019, mianiLOCO2022}. 

Conspiracy claims also frequently feature linguistic markers of uncertainty, such as the repeated use of exclamation marks and question marks \cite{mianiLOCO2022}. These features reflect their explanatory function---one that aligns with \citet{whitsonLacking2008}'s control theory, which posits that individuals may turn to conspiracy theories in response to environmental uncertainty, as a way of regaining perceived control. This tendency is further reinforced by the argumentative nature, which often challenges institutional narratives while highlighting gaps or inconsistencies in available evidence pieces \cite{oswaldConspiracy2016}. 

Social identities are often emphasised in conspiracy claims as well. They include pronominal polarisation (e.g., ``we'' vs. ``they'') or affiliative expressions (e.g., ``our,'' ``us,'' ``friend''), which emphasize in-group cohesion and out-group exclusion \cite{douglasUnderstanding2019}. Conspirators are often labelled as evil ``enemies'' who oppose and harm ``us'' to advance conspiratorial agendas \cite{lewandowskyConspiracy2020}. While prior research suggests that these psycholinguistic properties are predictive in the viral spread of conspiracy theories \cite{mianiLOCO2022}, it remains unclear how they correlate with persistence. We therefore propose the second research question: \textit{to what extent do the psycholinguistic changes affect the persistence of conspiracy claims online? (RQ2)}

Moreover, we investigated the changes of AAT categories in conspiracy claims, which captures the narrative-level granularity. Conspiracy narrative structures often consist of several basic elements including conspiratorial agents, their actions, plots, and the action's targets (i.e., goals, consequences, or victims) \cite{samoryGovernment2018}. They constitute a primary mechanism through which individuals form conspiratorial beliefs, connect seemingly unrelated events, and make sense of the world. 

A growing body of research employs AAT to study the diffusion of conspiracy theories. For example, through a decade-long analysis of Reddit conspiracy discussions, \citet{samoryGovernment2018} identified twelve recurrent AATs across diverse contexts. Prominent examples include the ``government agency, controls, communications'' triplets, where institutional groups manipulate information through concealment or distortion; the ``political leader, usurps, power'' triplets, portraying powerful figures seeking personal or political gain; and the ``organisation, pursues, profit'', framing institutions as threats to wellbeing. Additionally, AATs can function as knowledge graphs for detecting emerging conspiracy claims. For instance, \citet{chongrealtime2021} developed a platform to re-contextualise conspiracies by tracing AATs network. Collectively, these studies demonstrate that focusing on AATs is a appropriate strategy for studying narrative level mutation of conspiracy claims. 

Empirical evidence has shown that AAT configurations mutate, while the core assertions of long-lasting conspiracies often remain relatively stable. For example, in the ``Pizzagate'' conspiracy, \citet{tangherliniautomated2020} showed that various actants, including Hillary Clinton, the Clinton Foundation, and James Alefantis, were all linked to the same false allegations of child trafficking. A similar pattern was observed during the COVID-19 pandemic in relation to ``lab-leak'' conspiracies. Early claims focused on the Wuhan Institute of Virology in China \cite{Why2021}; later variants however implicated the Canadian Winnipeg laboratory \cite{paulsOnline2020}; and the U.S. Fort Detrick military biolab \cite{Wuhan2021}. The continual substitution of the \textit{target} element appears to have sustained the diffusion and survival of the lab-leak conspiracy theories.

Despite the suggestive pattern of a relationship between AAT mutations and conspiracy claim persistence, few studies have systematically examined the extent to which these mutations affect persistence. We therefore pose our third research question: \textit{to what extent does AAT category change affect the persistence of conspiracy claims online? (RQ3)}

While RQ1–RQ3 quantitatively assess the impact of language mutations at different analytical levels on persistence, they do not reveal how mutations unfold. To gain an in-depth understanding of language mutations in conspiracy claims, we conducted a qualitative analysis on a sample of long-lasting conspiracy claims, focusing on their interpretable changes in psycholinguistic properties and AAT categories. Accordingly, we propose our fourth research question: \textit{how do long-lasting conspiracy claims mutate in terms of psycholinguistic properties and AAT categories? (RQ4)} 

\section{Methods} %1000 words

\subsection{Data Collection}

We analysed the mutation and persistence of conspiracy claims using a three-year COVID-19 dataset (Jan 2020 --- Nov 2022) \path{COVID-19-TweetIDs} collected from Twitter,\footnote{Now known as X. We use the original name to reflect the period during which the data were collected.} \cite{chenTracking2020} which includes 2.77 billion related posts collected in real-time throughout the pandemic using streaming API. We retrieved tweet IDs and re-hydrated within one week of the original posting timestamp. This serves as our source dataset.

\begin{figure*}
    \centering
    \includegraphics[width=\textwidth]{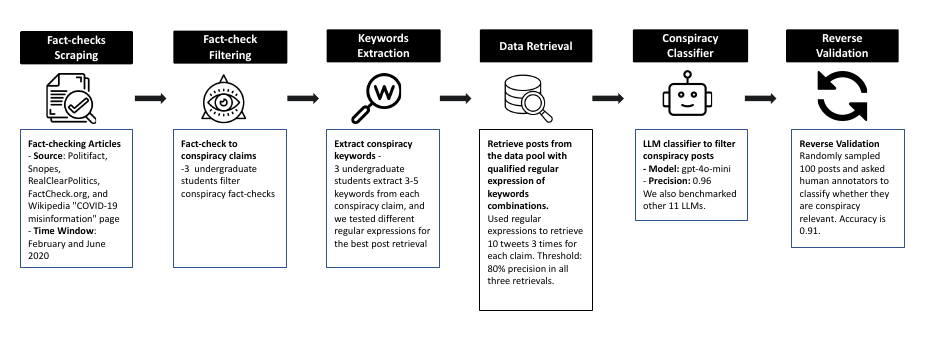}
    \caption{The workflow of data collection. }
    \label{fig:data}
\end{figure*}

We retrieved conspiracy posts from the source dataset as follows. As shown in Figure \ref{fig:data}, we first generated 37 regular expressions based a list of conspiracy keywords collected from 562 fact-checking articles.\footnote{Fact-checking articles were scraped between February and June in 2020 from Politifact, Snopes,RealClearPolitics, FactCheck.org and ``COVID-19 misinformation'' Wikipeida page \url{https://en.wikipedia.org/wiki/COVID-19_misinformation}.} Then we used keyword-matching to retrieve conspiracy posts from the source dataset. Next, to validate the relevance, ensuring data collected are truly conspiracy related, we filtered the dataset with a few-shot LLM classifier \path{GPT-4o-mini} ($F1 = 0.90$) where 22.21\% posts were removed.\footnote{We benchmarked a dozen of LLMs against a ground truth conspiracy dataset ($N = 150$) annotated by two postgraduates. \path{GPT-4o-mini} strikes a good balance between performance (i.e., precision) and cost. Please see the Appendix for more details regarding LLMs conspiracy classifier.}. Finally, we conducted a reverse validation where we randomly sampled 100 posts from the dataset and asked human annotators to label them using the same prompt to ensure that the dataset predominantly consists of conspiracy related discussions. Through this identification and validation process, 446,829 conspiracy posts were collected for analysis. The data is English only. %\footnote{Note the dataset does not only contain conspiracy endorsement, but also posts against or showing suspicion to conspriacy claims. }

\subsection{Conspiracy Claim Matching}
We examine the persistence at the \textit{claim} level of conspiracy theories. A claim refer to discrete statements or assertions that convey one conspiratorial idea. Different posts may refer to the same underlying conspiracy claim. Following prior research on fact-checking, we conceptualize this as a claim-matching task, in which semantically similar posts are clustered into the same conspiracy claim group. As a result, posts within the same claim group share a similar underlying assertion that can potentially be addressed by a single fact-checking claim \cite{kazemiClaim2021a}.

Operationally, we employed a threshold-based semantic similarity clustering method \cite{quelleLost2025}. First, all posts are embedded into a vector space using sentence embeddings. Next, pairwise semantic similarities are computed, and an edge is established between two posts if their similarity exceeds a predefined similarity threshold. This procedure constructs a similarity graph, from which claim clusters are identified as connected components.

We used the \path{all-mpnet-base-v2}\footnote{Accessed from sentence-transformer \url{https://huggingface.co/sentence-transformers/all-mpnet-base-v2}} embedding model, which is one of the best-performing sentence embedding models\footnote{See the model ranking on Sentence Transformers \url{https://www.sbert.net/docs/sentence_transformer/pretrained_models.html}} that has been widely used in clustering social media posts \cite[e.g.,][]{mansourDid2022, singhUTDRM2023}, and cluster similar posts with a cosine similarity exceeding a given threshold. Then at the pairwise semantic similarities computation step, we implemented Locality Sensitive Hashing (LSH) using the \path{ANNOY} library\footnote{Accessed from \url{https://github.com/spotify/annoy.}} to efficiently retrieve approximate nearest neighbours.  The retrieval algorithm begins with an initial set of 10 tweets and iteratively doubles the number of candidates until the cosine similarity of the most distant neighbour falls below a predefined threshold \cite{quelleLost2025}. This approach can efficiently reduce the computation load for large-scale dataset.

To determine the optimal threshold, we evaluated the quality of clusters by quantitatively examining the goodness of fit (i.e., the intra- and inter-cluster distances); and qualitatively validating the semantic coherence of the most dissimilar post pairs within each cluster. The optimal threshold should strike a balance between minimizing intra-cluster distance and maximizing inter-cluster distance. We settled at the threshold at 0.88, aligning with findings from previous research \cite{quelleLost2025, kazemiClaim2021a}. Details of optimal threshold selection is in the Appendix. 

\subsection{Psycholinguistic Property Measurement}

The psycholinguistic properties were measured using the fifth version of the Linguistic Inquiry and Word Count software (LIWC22) \cite{boydDevelopment2022}, a validated dictionary widely used to examine social-psychological processes in textual data \cite{chungWhat2018, tausczikPsychological2010}. LIWC22 contains over 12,000 words, word stems, and phrases, each assigned to one or more categories that correspond to specific psycho-social constructs. It is organized into 10 macro-categories and 118 subcategories, capturing varying levels of psycholinguistic granularity. Thus it offers one of the most comprehensive taxonomies of psycholinguistic constructs available, providing a robust framework for the analysis.

Next, we selected psycholinguistic categories that are known in predicting the viral diffusion of conspiracy claims or conceptually associated with conspiratorial thinking. To begin with, we drew on \citet{rainsPsycholinguistic2023} who examined the role of LIWC psycholinguistic categories in predicting virality. They identified 14 LIWC categories that significantly predicted the viral spread of conspiracy posts on Twitter. If a psycholinguistic feature is significantly associated with viral diffusion, it is also theoretically relevant to examine in relation to persistent diffusion. Therefore, we included them in our empirical test. 

Next, we qualitatively re-assessed the remaining LIWC categories based on their conceptual alignment with the definition of conspiracy theories. Following \citet{douglasWhat2023} \citet{vanprooijenPsychology2018}, and \citet{lewandowskyConspiracy2020}, we added 4 more LIWC categories: power, moral, conflict and cognitive processes. The LIWC \textit{power} category captures power relationships that are often central to conspiracy content; \textit{moral} relates to conspiratorial framing around threat and secrecy; and \textit{conflict} reflects intentionality and threat; and the \textit{cognitive processes} speaks to causal reasoning in conspiracy discourse. 

Finally, we merged some closely related sub-categories for clarity in the analysis: for example combining time-related subcategories (past, present, and future) into a single ``time orientation'' dimension. As a result, we came up with a ten-category framework for the mutation analysis. Table \ref{tab:liwc22} shows the detailed information of final selected categories and example vocabularies.\footnote{Please refer to the Appendix for detailed LIWC psycholinguistic category description.} 

\begin{table*}[t]
\caption{Psycholinguistic properties and corresponding LIWC22 categories with examples adapted from \citet{rainsPsycholinguistic2023}.}
\label{tab:liwc22}

\begin{tabular*}{\textwidth}{@{\extracolsep{\fill}} 
p{2.5cm} 
p{3cm} 
p{5cm} 
p{5cm} @{}}
\toprule
\textbf{LIWC22 Labels} & \textbf{Category} & \textbf{Description} & \textbf{Example Words} \\
\midrule

ppron & Personal Pronouns & Words referring to people in communication, including 1st, 2nd, 3rd persons & I, you, she, we, they \\

time\_orientation & Time Orientation & Words indicating when things happen, including subcategories time, focus past, focus present, and focus future & past, future, now, yesterday, soon \\

health & Health & Words showing physical and mental health terms, including subcategories illness, wellness, and mental health & medic*, patients, physician*, health \\

cogproc & Cognitive processes & Words related to thinking and reasoning activities, including subcategories insight, causation, discrepancy, tentative, certitude, differentiation & know, cause, understand, think, believe \\

tone\_neg & Negative Emotions & Words expressing negative feelings or emotional states & bad, hate, hurt, tired, awful \\

socrefs & Social Referents & Words referring to groups or social relationships, including subcategories family and friends, male and female references & parent*, mother*, father*, friend*, dude \\

conflict & Interpersonal conflict & Words describing referents to concepts indicative of or reflecting confrontations & fight, kill, killed, attack \\

moral & Moralization & Words reflecting judgmental language, where the speaker evaluates behavior or character & wrong, honor*, deserv*, judge \\

power & Power & Words reflecting ownership, power dynamics, or hierarchy & own, order, allow, power \\

risk & Risk & Words associated with danger or uncertainty & danger, threat, safety, warning, caution \\
\bottomrule
\end{tabular*}
\end{table*}

We define language mutations of psycholinguistic properties as the percentage change in the frequency of matched words associated with a given psycholinguistic category. Specifically, we computed the the absolute difference in LIWC22 category scores between two consecutive posts in the conspiracy claim, and considered it a mutation\footnote{For example, if one post contains 20\% \textit{anger} words and another contains 5\%, this is considered a significant mutation.} if the change exceeded a threshold of 50\%, \footnote{We also conducted a sensitivity test regarding the mutation threshold in the Appendix.} that is: $$\left| \frac{\text{liwc}_a - \text{liwc}_b}{(\text{liwc}_a +\text{liwc}_b)/2 } \right| >= 0.5$$ Moreover, we included the fluctuation of changes by measuring the standard deviation of percentage difference within clusters. 

\subsection{AAT Measurement}
We performed a LLMs-assisted AAT extraction. First, we preprocessed the posts to reduce noise by removing URLs, mentions, HTML tags, emojis, and by converting text to lowercase. We also lemmatized forms of words such as plural nouns and verb tenses \cite{samoryGovernment2018}. 

Second, we used large language models (LLMs) to extract AATs from the preprocessed posts. Prior research has shown that LLMs can extract AATs from natural language with excellent accuracy \cite{salmanTiny2024}. We thus tested three models: \path{en-core-web-trf}, \path{llama3.3-70b-instruct}, and \path{gpt-4o-mini}, on a sample of manually annotated conspiracy posts ($N =20$). Model performance was evaluated on accuracy, meaningfulness, and computational efficiency. As a result, \path{gpt-4o-mini} significantly outperformed other models. We further tuned prompts following \citet{neubergerUniversal2024}'s universal prompting strategy.\footnote{See Appendix for more details on the prompt design and experimental results.}

Third, we clustered actors, actions, and targets separately to form meaningful groups using K-Means, based on \path{text-embedding-3-large} embedding model.\footnote{We compared \path{text-embedding-3-large} with \path{all-mpnet-base-v2} used in the claim clustering task. The former produced superior clustering results on short phrases, so it was selected for the AAT clustering task.} Before clustering, we filtered AATs by removing actors and targets expressed solely as pronouns, which are uninformative for capturing changes. We also removed targets that were solely verbs, as these typically correspond to subordinate clauses \cite{samoryConspiracies2018}. To determine the optimal $k$, we used the elbow method and silhouette scores, settling on $110$, $130$, and $95$ clusters for actor, action, and target groups, respectively.\footnote{Details of the optimal $k$ selection process are provided in the Appendix.}

Fourth, we used \path{gpt-4o} to label the resulting clusters. For each actor, action, and target cluster, we applied maximum marginal relevance (MMR) sampling to select a representative set of items for description. MMR balances relevance and novelty and fits well with K-Means clustering, while also reducing redundancy in the examples shown to the LLM. Using the centroid as the initial query, we sampled 20 items and asked the LLM to describe what they have in common. We then qualitatively validated and refined these descriptions. The prompt and example clusters are provided in the Appendix.

Mutation in AATs is defined as a change in the actor, action, or target clusters within a conspiracy claim. The following example from the dataset illustrates this definition. Within the same claim about child trafficking, the actor \textit{national media} in Tweet A is clustered with other similar actors in the ``media organisation'' group, whereas the actor in Tweet B \textit{democratic party} is clustered in the ``political party'' group. Since ``media organisation'' and ``political party'' represent different actor groups, this instance is counted as an actor mutation within the claim. We then applied survival models to estimate the impact of such mutations on the lifespan of conspiracy claims.

\begin{quote}
    Tweet A: ... \textit{national media} pretend child sex trafficking does not exits... \par
    Tweet B: ...\textit{democratic party} using child trafficking as a cover...
\end{quote}

\subsection{Qualitative Analytical Framework}

To assess how psycholinguistic properties and AATs mutate, we first coded observed changes systematically into three general mutation patterns: levelling, sharpening, and assimilation \cite{allportAnalysis1946}. Then we applied \citet{campbellSystematic1958}'s typologies to interpret the changes and to identify the mechanisms underlying the mutations. For this analysis, we purposefully sampled conspiracy claims with identifiable and interpretable mutations that persisted for more than seven days, and compared between posts within each claim.

For mutations in psycholinguistic properties, we randomly sampled 10 claims from long-lasting ones that include mutation of identified psycholinguistic properties in Table \ref{tab:liwc22} ($N=100$). Then we qualitatively coded the categories changes by comparing pair of posts in conspiracy claims. For mutations in AAT categories, similarly, we randomly sampled 10 claims and qualitatively coded the mutation patterns from sampled pairs. 

\subsection{Survival Analysis}

To examine the relationship between claim mutations and persistence, we applied survival analysis. We first used a non-parametric Kaplan–Meier (KM) model to explore group differences between mutated and non-mutated claims, in order to descriptively assess whether the lifespan of mutation group is significantly distinct from the non-mutation group. The event was defined as the absence of any conspiracy post within a claim cluster for thirty consecutive days. In other words, if a conspiracy claim did not appear in any tweets for thirty days, it was considered to have ended, as it was no longer actively discussed. The lifespan of a claim was therefore calculated as the time difference between the first and last post in the cluster.

The survival function is described below, where $S(t)$ represents the survival probability at time $t$, ${d_i}$ is the number of events (i.e., ``death'') at time $t_i$, and ${n_i}$ is the number of conspiracy claims at risk just before time $t_i$. 

\begin{quote}
    \centering$\hat{S}(t) = \prod_{t_i < t} \frac{n_we - d_i}{n_i}$ 
\end{quote}

Next, we applied an accelerated failure time model (AFT) to estimate the effect of semantic, psycholinguistic and AAT category mutations on claim persistence. We modelled the persistent diffusion as a Weibull distribution because of the temporal pattern of information on social media - information generally has a high hazard rate at the outset of transmission \cite{pfefferHalfLife2023, haleAnalyzing2024}. Thus, we assume a decelerating hazard for conspiracy claims: the risk of event is highest initially and decreases over time as the content continues to circulate. As a result, AFT Weibull model is sufficient to model such scenarios.

The survival function for Weibull distribution is defined as follows, where $\lambda$ represents the scale parameter (associated with covariate effects) and $\rho$ controls the shape of the hazard function. $\rho$ is less than 1 (concave cumulated hazard) in this study, indicating a decreasing hazard rate over time:

\begin{quote}
    \centering $H(t) = \left(\frac{t}{\lambda(x)} \right)^{\rho(y)}$
\end{quote}

For RQ1, we examined to what extent do early semantic mutations predict claims' persistence. Prior research suggests that about 95\% of social media content loses engagements within the first 24 hours after posting \cite{pfefferHalfLife2023}. Therefore, we defined the early mutation window as 24 hours following the initial post of a claim.\footnote{As a sensitivity test, we also examined a one-hour early window. The results remain consistent and are reported in the Appendix.} Specifically, we measured the semantic drift within this 24-hour window and modelled if these drifts predict later persistence. The analysis thus focuses on claims that survived beyond the first day. Semantic drift is defined as cumulative drift --- the average semantic distance of each early post within the claim and the earliest (seed) post. We additionally controlled for early post volume, number of likes, number of retweets, number of unique users, and users’ follower counts that might be correlated to the viral diffusion of posts. 

For RQ2, LIWC property mutation was operationalized both as a binary indicator and as the magnitude of fluctuation. AAT category mutation was similarly measured as a binary variable. Full model specifications and control variables are provided in the Appendix.

\section{Results} %2000 words

\subsection{Semantic Mutation and Persistence}

To answer RQ1, we first descriptively inspected the correlations between semantic mutations and claim persistence. We started by assessing whether conspiracy claims undergo semantic change over time by computing the average cosine similarity between indirectly connected post pairs within each claim. Under our clustering method, directly connected post pairs always exceed the preset threshold, while indirectly connected nodes may exhibit lower pairwise similarities. Plotting this similarity as a function of time difference between post pairs therefore provides an effective measure of semantic drift over claim's lifespan - if claim mutates, similarity should decline as the time gap between posts increases. 

\begin{figure*}[!htbp]
    \centering
    \includegraphics[width=0.9\textwidth]{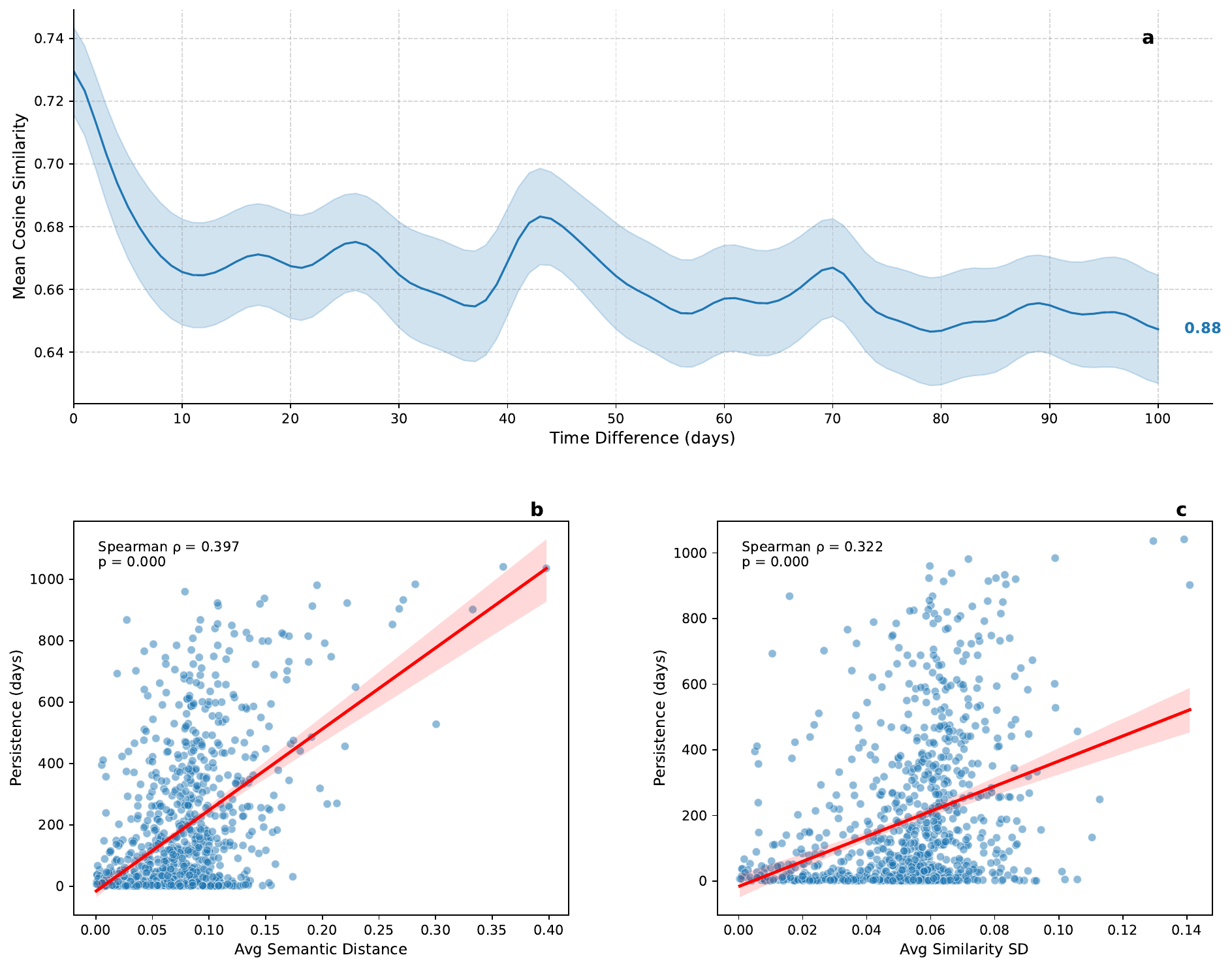}
    \caption[Cosine similarity and lifespan of conspiracy claims]{Plot \textbf{a} shows average cosine similarity of indirectly connected conspiracy posts over a hundred days. Time measures the number of days between the dates of two clustered posts not directly connected. Shading shows the standard error. Panel \textbf{b, c} show the relationship between claim lifespan and their average cosine similarity and standard deviation given the preset threshold of 0.88. The standard deviation (SD) quantifies how dispersed the cluster is relative to the mean cosine similarity.}
    \label{fig:cos_sim_rq1}
\end{figure*}

As shown in Figure \ref{fig:cos_sim_rq1} Panel \textbf{a}, we observed a consistent decrease in cosine similarity as time difference increased within the claim. The semantic similarity of indirectly connected nodes declined over 100 days, indicating that conspiracy claims undergo semantic mutations over time. 

Next, we examined whether the extent of semantic mutation is associated with claim persistence. As shown in Figure \ref{fig:cos_sim_rq1} Panel \textbf{b} and \textbf{c}, both the average semantic similarity and its standard deviation have shown strong positive correlations with claim persistence. 

While suggestive, these descriptive associations are limited in explaining the causal effect. Therefore, we turned to a Weibull AFT model, isolating the effect of early semantic mutations --- within the first 24 hours --- on subsequent claims' lifespan, and controlling for post counts, engagements and selected user information. 

\begin{figure}[!htbp]
    \centering
    \includegraphics[width=0.9\columnwidth]{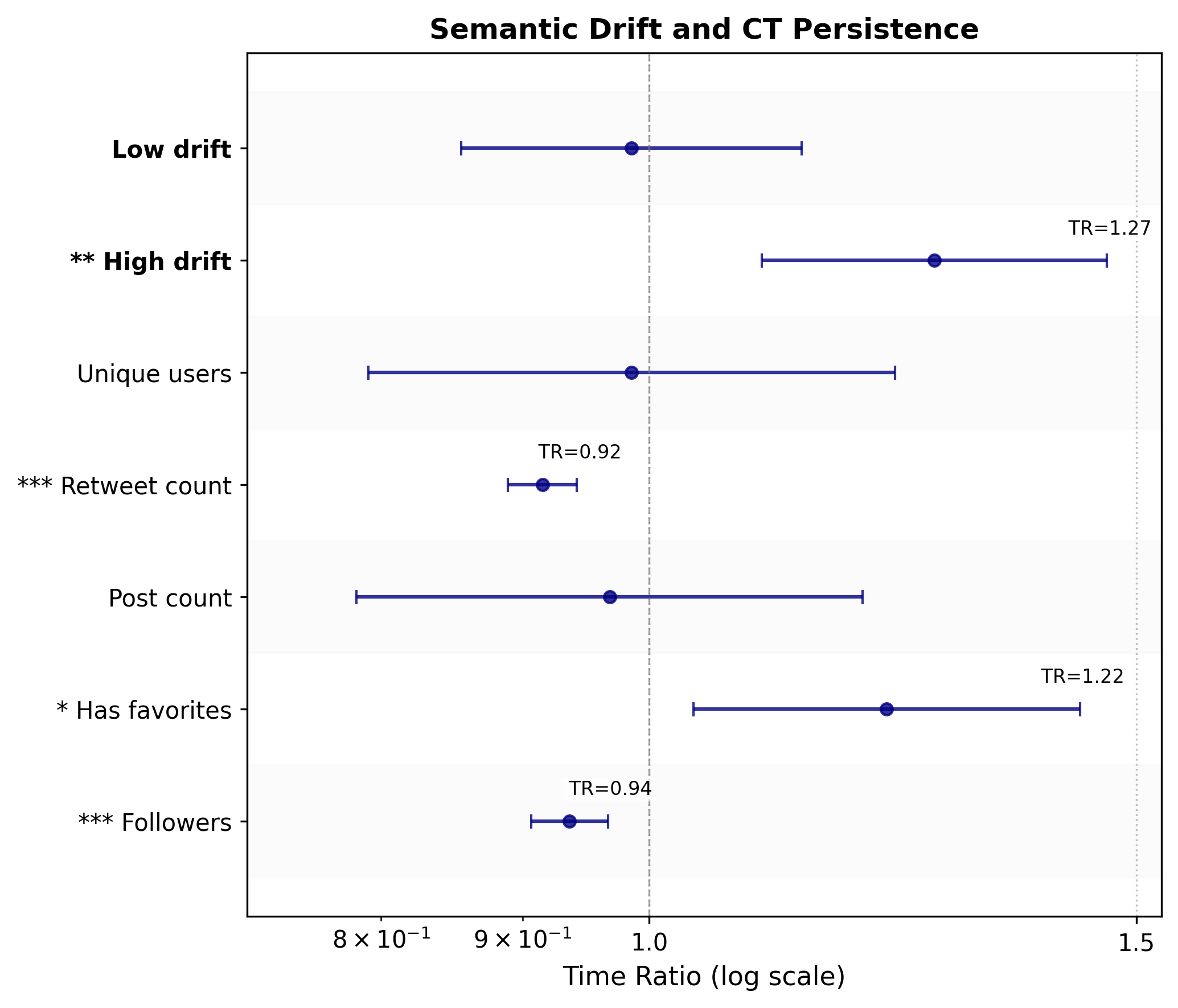}
    \caption[Weibull AFT model time ratios for early semantic drifts.]{The time ratios (TR) of semantic drift in Weibull Accelerated Failure Time (AFT) model. The x-axis represents the TR with 95\% confidence intervals, y-axis corresponds to covariates in the Weibull AFT model. Asterisk represents the significance of variables (* $p < 0.05$, ** $p < 0.01$, *** $p < 0.001$). The horizontal reference line indicates no effect. $TR > 1$ represents longer survival times (decelerated failure), while $TR < 1$ represents shorter survival times (accelerated failure). Focal variables are the semantic drift, bolded in text. Others are control variables in the model. Semantic drift is the cumulative cosine distance from the earliest post, averaged across all posts within the 24-hour early window in the claim. Claims are categorised into three groups: no drift (drift = 0), low drift (non-zero drift below the median), and high drift (non-zero drift above the median).  Median cosine distance = $0.006$. No drift serves as the reference category.%\footnote{The median split was chosen because the non-zero drift distribution is heavily right-skewed, rendering tertile boundaries semantically indistinguishable at the lower end.}. 
    }
    \label{fig:q1_sem_drift}
\end{figure}

As shown in Figure \ref{fig:q1_sem_drift}, relative to claims that spread without semantic drift, claims with early semantic drift survived 27\% longer ($TR = 1.27, p < 0.005$). Conspiracy claims with little drift where the accumulative semantic similarity is below the median threshold showed no significant persistent advantage ($TR = 0.99, p = 0.84$). %Notably, the median threshold separating low from high semantic drift in our sample was relatively small in absolute terms ($0.006$), suggesting that meaningful semantic drifts that extend claim lifespans are relatively easy to achieve in reality.

\subsection{Psycholinguistic Property Mutation}

\begin{figure*}[!htbp]
    \centering
    \includegraphics[width=0.9\textwidth]{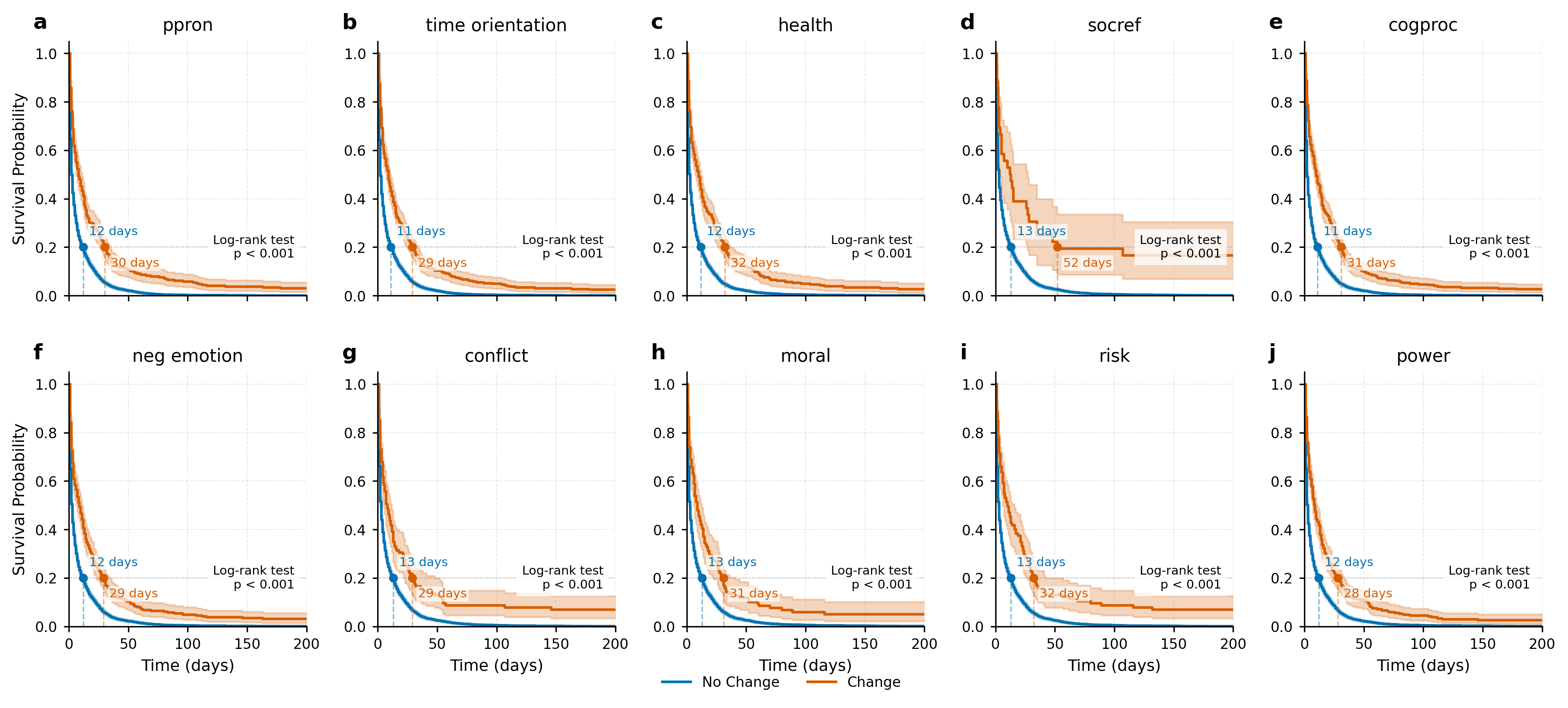}
    \caption[Kaplan–Meier survival curves for psycholinguistic properties]{Kaplan-Meier survival curves for psycholinguistic properties. The x-axis is days, and the y-axis represents the probability a claim is still transmitting after $t$ days. The vertical dash lines indicate the 20\% survival time for both groups. The log-rank tests shows the statistic difference between two curves in each plot.}
    \label{fig:q2_km}
\end{figure*}

To address RQ2, we applied similar analysis pipeline to the mutation of psycholinguistic properties of conspiracy claims. We first applied the KM models to describe the group difference. As shown in Figure \ref{fig:q2_km}, conspiracy claims with psycholinguistic property mutations (yellow) consistently displayed twice longer lifespan. For instance, conspiracy claims with changes in ``health'' categories (Figure \ref{fig:q2_km}-c) that express physical and health terms including illness, wellness, mental health and other body and mental related functions, exhibited longer lifespans than those without changes. The 20\% survival time was 33 days for mutated groups versus 11 days for non-mutated groups. Similarly, psycholinguistic elements such as person pronouns, time orientations, cognitive processes, moral, risks and power demonstrated 2 to 3 times longer lifespan than their non-mutated groups. 

% Similarly, as shown in Figure \ref{fig:q2_km}-h, conspiracy claims with mutations in ``risk'' words demonstrated 2.58 times higher 20\% survival time than non-mutate claims (31 versus 12 days). As the ``risk'' category estimate words associated with danger, threats and uncertainty (e.g., secur*, protect*, pain, and risk*), 
    
\begin{figure}[!htbp]
    \centering
    \includegraphics[width=0.9\columnwidth]{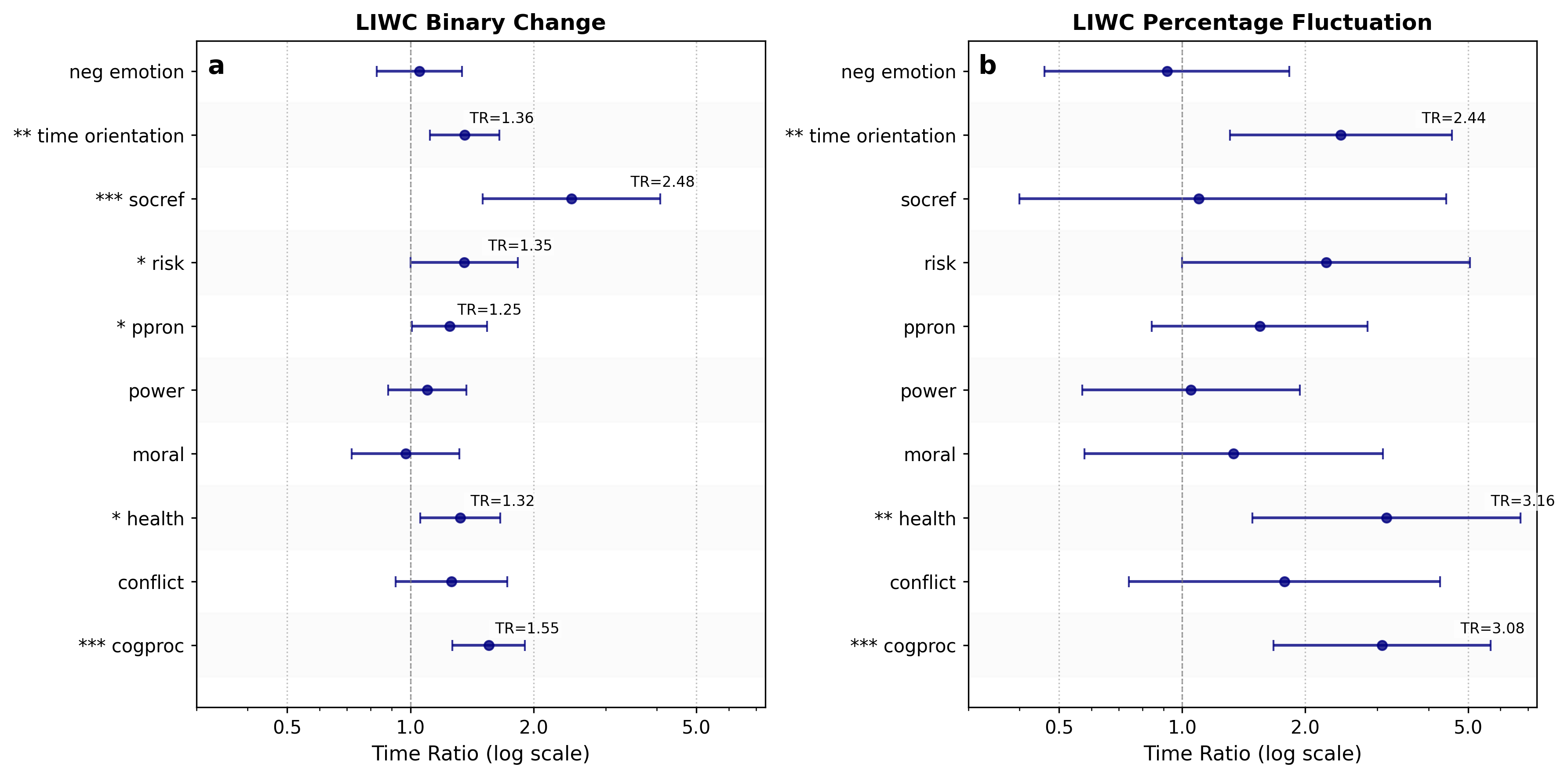}
    \caption[Weibull AFT model time ratios for psycholinguistic properties]{The time ratios (TR) of psycholinguistic properties in Weibull Accelerated Failure Time (AFT) model. Plot a shows the binary changes, and Plot b illustrates the standard deviation of percentage change in conspiracy claims. The x-axis represents the TR with 95\% confidence intervals, y-axis corresponds to covariates in the Weibull AFT model. Asterisk represents the significance of variables (* $p < 0.05$, ** $p < 0.01$, *** $p < 0.001$). The horizontal reference line indicates no effect. $TR > 1$ represents longer survival times (decelerated failure), while $TR < 1$ represents shorter survival times (accelerated failure).}
    \label{fig:q2_aft}
\end{figure}

Next, we applied the Weibull AFT model to estimate the effect of psycholinguistic mutations on the survival time of conspiracy claims. Figure \ref{fig:q2_aft}-a shows the time ratio (TR) of mutation factors (binary). Mutations in social reference words ($TR=2.22, p<0.001$) has the most significant TR where conspiracy claims with changes in family and friends words survive 122\% longer than their non-mutated counterparts. Mutations in cognitive process terminology ($TR=1.67, p<0.001$) extend survival by 67\%. Similar statistically significant effects are also observed in time orientation words ($TR=1.30, p<0.01$), risk words ($TR=1.43, p<0.05$), and health words ($TR=1.30, p<0.05$). 

Figure \ref{fig:q2_aft}-b further estimates the effect of the extent of changes in addition to the presence of changes. Here, health related words show the most significant impact ($TR=2.74, p<0.001$), with claims experiencing one more unit of fluctuation (i.e., standard deviation of LIWC percentage differences among posts) change in health language surviving almost three times longer. Similarly, cognitive process fluctuations significantly increase survival time ($TR=2.57, p<0.001$), as do changes in time orientation language ($TR=1.78, p<0.01$). It suggests that the magnitude of changes in health, cognitive processing, and time psycholinguistic categories are significant in predicting the persistence of claims as well.

\subsection{AAT Categories Mutation}

\begin{figure}[!htbp]
    \centering
    \includegraphics[width=0.9\columnwidth]{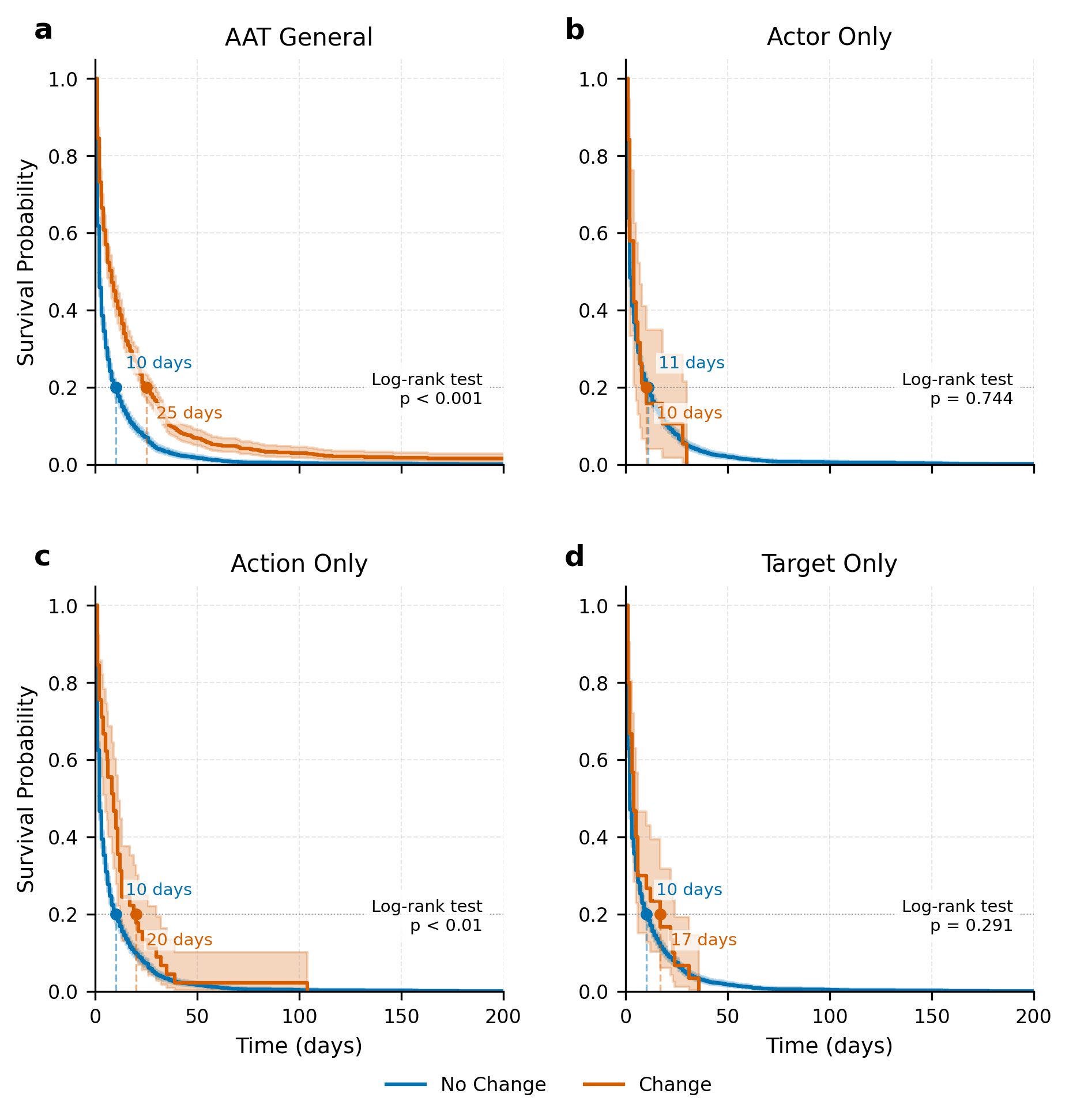}
    \caption[Kaplan–Meier survival curves for AATs]{Kaplan-Meier survival curves for AAT properties. }
    \label{fig:rq3_aat_km}
\end{figure}

We applied the same analysis pipeline to investigate mutations in AAT categories. As shown in Figure \ref{fig:rq3_aat_km}, conspiracy claims with AAT category mutations consistently live longer than their unchanged counterparts, with 20\% survival time of 25 days compared to just 10 days for unchanged ones. For individual AAT components, we observed similar patterns in action and target only groups. Action-only and target-only mutations (Figure \ref{fig:rq3_aat_km}-c and -d) demonstrated identical survival patterns with 20\% survival times of 20 and 17 days compared to 10 days for their unchanged counterparts. On the other hand, actor-only changes (Figure \ref{fig:rq3_aat_km}-b) is not informative due to limited data points. 

Weibull AFT model was again applied to estimate the effect size. As shown in Table \ref{tab:aat_aft}, changes in actor, action, and target all have substantial positive associations with the lifespan of conspiracy claims. When examined individually (Models 1-3), each type of change shows a significant similar effect, approximately tripling the expected survival time (acceleration factors of 2.91, 2.88, and 2.86 respectively). In the multivariate model, action change emerges as the strongest independent predictor, increasing claim lifespan by 68\% ($TR = 1.68, p < 0.001$), while actor change increases lifespan by 46\% ($TR = 1.46, p < 0.01$) and target change increases lifespan by 40\% ($TR = 1.40, p < 0.05$). Also, we preformed the interaction (Model 5), which shows that only action change maintains a strong independent effect, increasing claim lifespan by 88\% ($TR = 1.88, p < 0.01$) when controlling for all other AAT factors and interactions. It suggests that mutations in action, actor, or target all appear to extend the lifespan of conspiracy claims, with action category mutations displayed the most significant effect. 
\begin{table}[t]
\small
\setlength{\tabcolsep}{1pt} % reduce column spacing

\caption{Weibull AFT models for conspiracy claims longevity. Exponentiated coefficients shown with standard errors in parentheses. $^* p < 0.05$, $^{**} p < 0.01$, $^{***} p < 0.001$.}
\label{tab:aat_aft}
\begin{tabular*}{\columnwidth}{@{\extracolsep{\fill}} lccccc @{}}
\toprule
\textbf{Variable} & \textbf{M1} & \textbf{M2} & \textbf{M3} & \textbf{M4} & \textbf{M5} \\
\midrule

\multicolumn{6}{l}{$\lambda$ (exp(coef))} \\
Intercept & 6.23$^{***}$ & 5.87$^{***}$ & 5.99$^{***}$ & 5.75$^{***}$ & 5.81$^{***}$ \\
A1 & 2.92$^{***}$ & --- & --- & 1.46$^{**}$ & 1.07 \\
A2 & --- & 2.89$^{***}$ & --- & 1.68$^{***}$ & 1.88$^{**}$ \\
T & --- & --- & 2.86$^{***}$ & 1.39$^{*}$ & 1.26 \\
A1 $\times$ A2 & --- & --- & --- & --- & 1.01 \\
A1 $\times$ T & --- & --- & --- & --- & 1.79 \\
A2 $\times$ T& --- & --- & --- & --- & 0.93 \\
A1 $\times$ A2 $\times$ T & --- & --- & --- & --- & 0.83 \\
\midrule
\multicolumn{6}{l}{\textbf{Model Statistics}} \\
Intercept $\rho$ & 0.72$^{***}$ & 0.72$^{***}$ & 0.72$^{***}$ & 0.73$^{***}$ & 0.73$^{***}$ \\
% Observations & 2374 & 2374 & 2374 & 2374 & 2374 \\
% Events & 2374 & 2374 & 2374 & 2374 & 2374 \\
% Log-likelihood & -7484.31 & -7464.31 & -7470.11 & -7452.10 & -7450.86 \\
% AIC & 14974.62 & 14934.62 & 14946.21 & 14914.19 & 14919.72 \\
Concordance & 0.55 & 0.56 & 0.56 & 0.57 & 0.57 \\
LR test $\chi^2$ (df) & 208.71 & 248.71 & 237.12 & 273.14 & 275.61 \\

\bottomrule
\end{tabular*}
\end{table}

\subsection{Qualitative Analysis on Mutation Patterns}

\begin{figure*}[!htbp]
    \centering
    \includegraphics[width=0.9\textwidth]{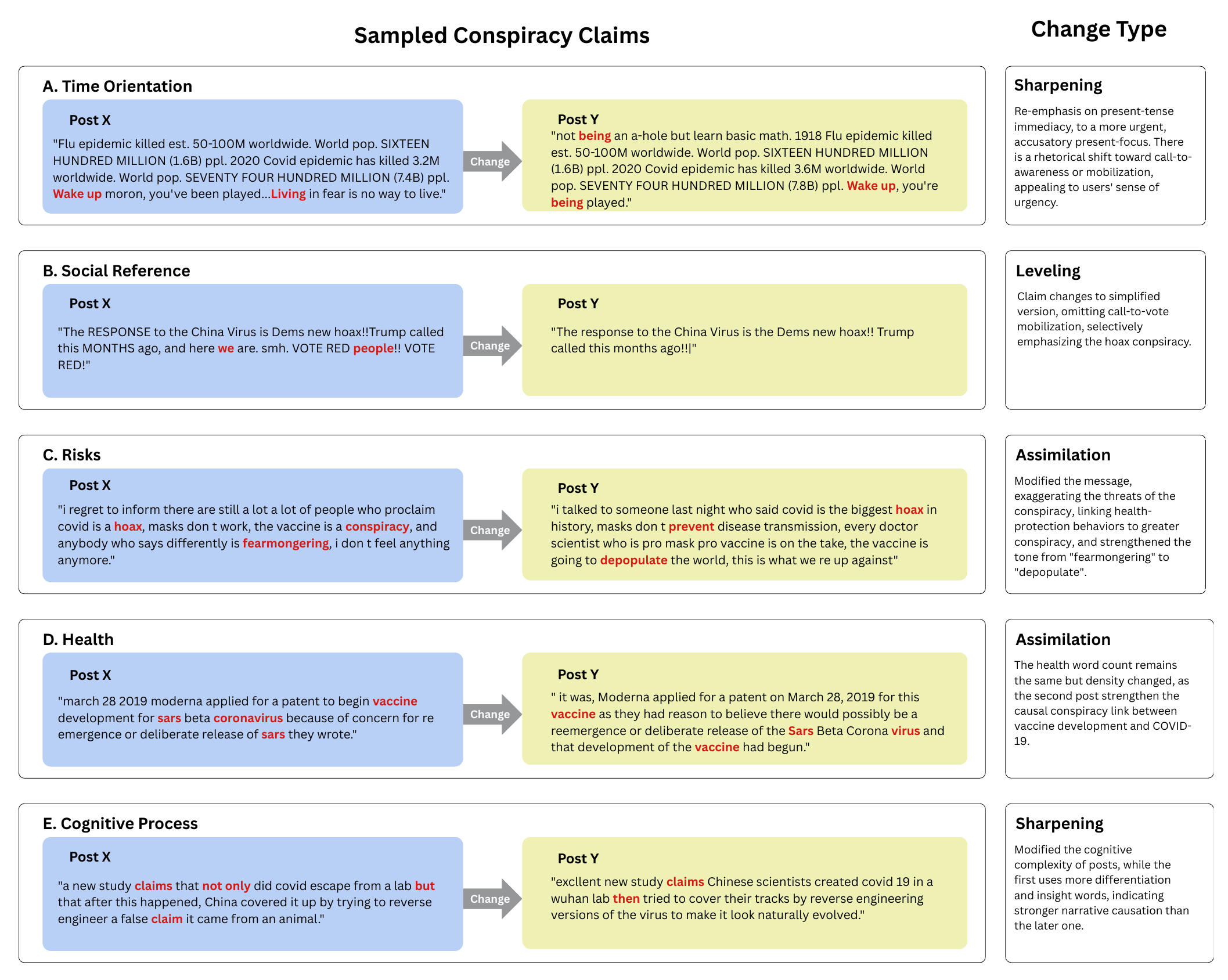}
    \caption[Qualitative analysis of psycholinguistic mutations]{Qualitative analysis of significant psycholinguistic elements change within conspiracy relevant claims.}
    \label{fig:q2_quali}
\end{figure*}

To gain an in-depth understanding of how psycholinguistic properties and AAT categories change within conspiracy claims, we conducted qualitative analysis at the lexicon level. For psycholinguistic properties, we purposefully sampled conspiracy claims that showed significant changes in time orientation, social reference, risks, health and cognitive process, and qualitatively analyzed their mutation patterns. Figure \ref{fig:q2_quali} shows examples of how psycholinguistic words shift. We observed the following two main patterns in the sampled data.

Simplification. We observed that the simplification in conspiracy claim changes can lead to three directions. Above all, labeling. Instead of scrutinized the conspiracy claims, users tend to simply stigmatize the event or figures and label it as ``hoax'' or other illegitimatized words (example B). Secondly, call for actions. Users' often react to conspiracy claims with simple catchy appeals such as calling for waking up (example A), voting (example B), and against institutions (example C). \footnote{Similar patterns were also widely observed in other conspiracy claims \cite[e.g.,][]{jeppesenCapitol2022}.} Thirdly, reduce the cognitive complexity. As indicated in example E, users might use fewer words that refer to thinking, memory, and reasoning, simplifying the cognitive process. The simplification often results in emphasizing on causation words, sharpening the accusation dimension. 

The other pattern is assimilation. The most salient assimilation types is ``assimilation to own attitudes'' --- where humans are prone to bias in the direction of their own opinion \cite{campbellSystematic1958}. For instance, as shown in example C, the language in post X becomes amplified and more extreme in the subsequent post where the language shifting from reporting others' beliefs (``people who proclaim'') to presenting more definitive conspiratorial claims (``biggest hoax in history'') and introducing additional conspiracy elements like ``depopulate the world'' which implies their own attitudes. 

\begin{figure*}[!htbp]
    \centering
    \includegraphics[width=0.9\textwidth]{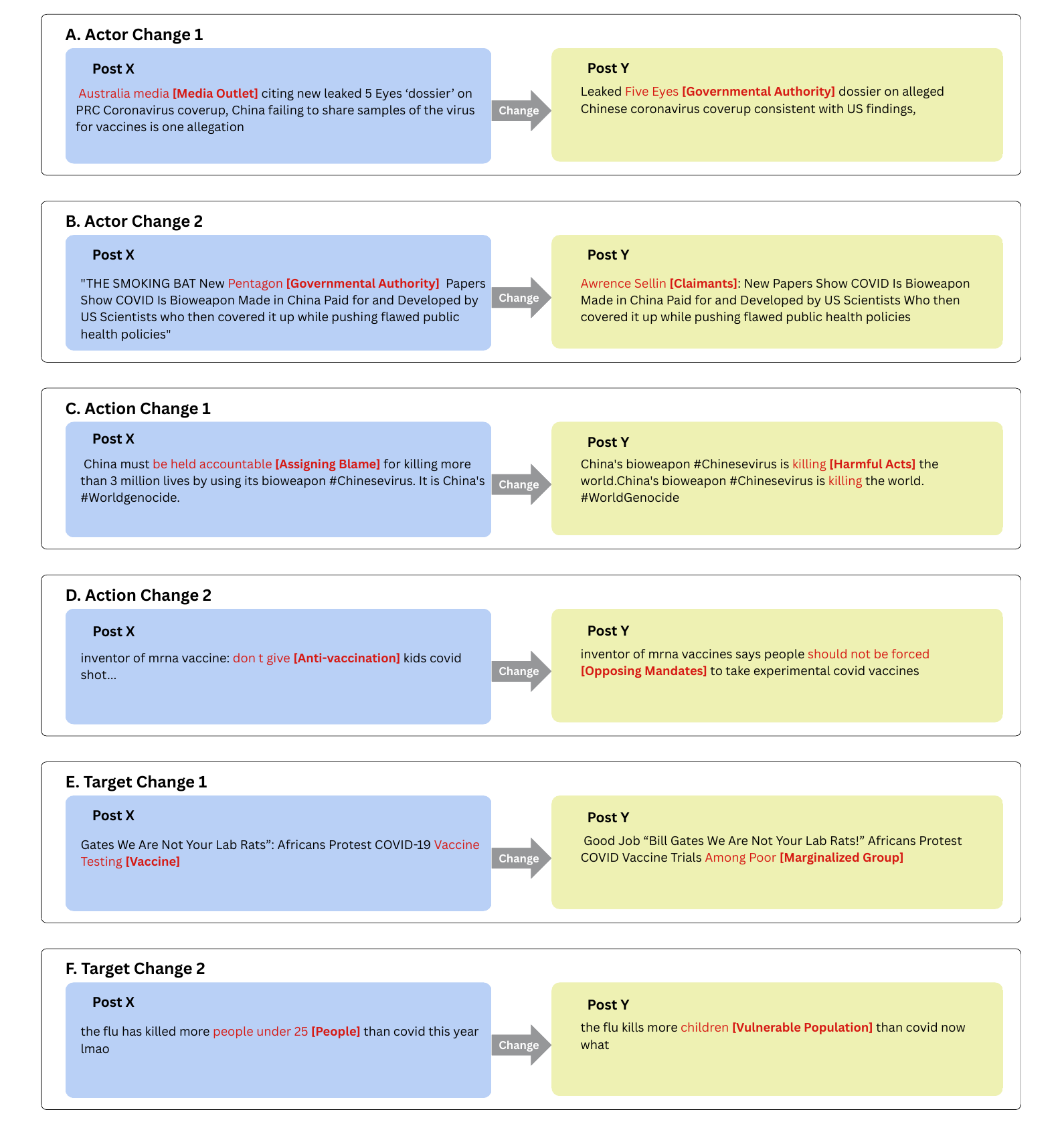}
    \caption[Qualitative analysis of AAT mutations]{Qualitative analysis of significant actor-action-target elements change within conspiracy relevant claims.}
    \label{fig:q3_quali}
\end{figure*}

We conducted similar qualitative analysis of sampled conspiracy claims to gain in-depth understanding of how AATs mutated. Figure \ref{fig:q3_quali} shows the examples of AAT category mutations. ``Government authority'' is among the most common actor changes categories. As shown in example A and B, ``government authority'' often intertwine with ``media outlet'' and individual ``claimants'' in conspiracy claims. In the action category, for similar conspiracy claims, the action may shift foci, for example from ``assigning blames'' to ``harmful acts'' (example C); or from ``anti-vaccination'' appeals to more general concerns about authoritative forced behaviours ``opposing mandate''. Among the target category, the most significant observation is the involvement of marginalized and vulnerable groups. As in example E, claim mutated from vaccine ``testing'' to testing on ``poor'' (people). Also example F shows that the victims changed from young people to ``children'', underlining the evolvement toward targeting vulnerable populations, potentially amplifying emotional appeal and perceived threat severity. \footnote{Note that the identified changes are not necessarily directional. They are sampled pairs of posts at two discrete time points.}

\section{Discussions} 

This study finds that language mutations in conspiracy claims are significantly associated with persistent diffusion. Specifically, conspiracy claims with early semantic drift have 27\% higher chance to survive longer online. It provides the first empirical evidence quantifying the effect of semantic drift on the lifespan of conspiracy claims. They extend the language mutation theories by incorporating a longitudinal perspective. The relatively small median threshold suggests that, conspiracy claim does not need a massive rewrite to survive longer. Even small semantic tweaks, such as swapping cognitive process words, or slightly altering a social reference, which are projected as little drift in the vector space, are sufficient to sustain the claim's lifespan. 

We also examine the effect of psycholinguistic properties and AATs categories mutations. Claims with psycholinguistic property mutations exhibit longer lifespans, where mutations in social reference, time orientation, risk, health, and cognitive process words had significant predictive power. Qualitative analysis further identifies two prevalent patterns of this micro-level conspiracy mutation: simplification and assimilation. In these cases, users tended to simplify linguistic features or incorporate their own personal opinions when altering conspiracy content. The simplification mutation pattern may be partly attributable to people's limited cognitive capacity to process complex information \cite{fiskeSocial2013}. Individuals often rely on cognitive shortcuts, which can lead to ``reductive coding,'' where inputs are simplified into less complex outputs \cite{campanFighting2017}. Since conspiracy claims are often perceived as more complex than non-conspiratorial explanations \cite{marshCompelling2022}, human actors may systematically introduce simplification mutations in conspiracy transmission.

The findings on the simplification mutation pattern inform future content moderation strategies for debunking the longitudinal diffusion conspiracy theories. Since mutations often follow a simplification pattern, tracking and debunking efforts should prioritise core, underlying claims rather than comprehensive narratives. That is, to account for the variation of conspiracy of claims in transmission, one potential effective approach is to decompose input into standardized unit of claims, such as atomic claims. Recent research in AI-assisted fact-checking has shown that prioritising atomic claims --- units of information that convey a single, simplified idea --- as the basis for interventions can improve the effectiveness of online content moderation \cite[e.g.,][]{wrightGenerating2022,metropolitanskyEffective2025}.  Future research should explore more on how atomic-claim-based debunking can be integrated into automated detection and debunking systems to address evolving variants of conspiracy theories.

The assimilation mutation pattern reflects another common human error: distorting messages toward the transmitter’s own attitudes. Due to the associative memory effect \cite[e.g.,][]{campbellIndirect1950, hymanInterviewing1954}, people are prone to bias outputs away from inputs in order to align with pre-existing opinions and beliefs. Findings on assimilation in conspiracy claims could also inform debunking strategies that incorporate public opinion monitoring. For example, survey data on attitudes toward particular conspiracy topics could be used to predict potential mutation directions and thereby support more effective moderation of conspiracy theories online. 

In addition, this mutation pattern may facilitate the design of user-customised debunking strategies. Rather than directly correcting conspiracy beliefs, future research should explore AI agent interventions that employ empathic, evidence-based reasoning to engage users in personalised dialogues. Such approaches acknowledge users’ specific concerns while gently introducing counter-evidence, which has already show durably correction effect to combat conspiracy persistence \cite{costelloDurably2024a}. By aligning corrective efforts with individuals’ existing cognitive frameworks, emotional needs, and ideological orientations, this strategy may reduce defensive reactions when beliefs are directly challenged, thereby being more effective in misbelief corrections.

The findings on AATs mutations also highlight several points. First, institutional entities account for a prominent proportion of actor groups. Among the top ten actor groups, more than half are institutional entities (e.g., Chinese authorities, government bodies, health professionals, health institutions, democratic actors, and political actors).\footnote{Table \ref{tab:aat_top10} displays the top ten AAT groups from the clustering results.} This aligns with prior research suggesting that many conspiracy theories are underpinned by distrust in authorities and institutional knowledge \cite{brashierConspiracy2023}. These actors often link into other categories such as ``media outlet,'' manifesting potential conspiratorial coalitions and further delegitimising institutional knowledge.

Second, we find that the most frequent actions in conspiracy claims involved accusations of harmful behaviours or activities related to information seeking (see Table \ref{tab:aat_top10}). This suggests that many conspiracy claims derive from suspicions of ``hidden information'' and ``concealment,'' which may hinder individuals’ ability to ``seek information'' or ``explain ideas.'' Information seeking is thus a primary motivation for people to seek conspiracy theories when uncertainty is high. This underscores the importance of transparent communication in times of crisis, which plays a crucial role in preventing individuals from turning to conspiratorial explanations. 

Finally, the qualitative analysis of target groups reveals frequent references to vulnerable populations, including the poor, children, women, and other marginalised communities. This resonates with previous research on the imbalanced power relations in claims, suggesting that conspiracies may function not only as illegitimate knowledge but also as expressions of deeper social conflicts \cite[e.g.,][]{fensterConspiracy2008, prattTheorizing2003, watersConspiracy1997}. Future studies should investigate the vulnerable framing in conspiracy claims and how it may affect their diffusion and persuasiveness in communication. 

\section{Limitations and Conclusions}
Nevertheless, Several limitations should be acknowledged in this study. Above all, this paper only focuses on the persistent diffusion at the claim level, which does not account for mutation across claims (i.e., cases in which a conspiracy claim mutates too much into a substantively different claim). Future research should explore more on the topical level mutation and further understand how conspiracy narratives evolves online in a more coarse granularity.

Next, LIWC is not able to capture specific word mutations. As LIWC only measures word percentage of certain psycholinguistic category, it does not quantify lexical changes within certain psycholinguistic category. For example, if two posts used different ``angry'' words with the same proportion in the same claim, it is not counted as a mutation in the dataset. Consider Post A: ``President Trump is determined to \textit{fight} deep state'' and Post B: ``Trump: we will \textit{demolish} the deep state.'' They used different ``anger'' words in the same claim, it is supposed to be a case of mutations in ``anger'' psycholinguistic category. However, the LIWC only measured the proportion of ``anger'' word --- 12.50\% and 14.28\% for post A and B respectively, thus no mutation will be measured. In other words, the mutation of psycholinguistic properties might be underestimated in this work. 

Similarly, AAT mutations also represent a conservative estimation. As shown in Figure \ref{fig:kmeans_vis}, K-Means clusters are not always coherent where AATs are not densely connected. Dissimilar words might be clustered into the same cluster. %For instance, ``violent shootings, stabbings, burnings'' and ``alien invasion'' belong to the same action group ``Harmful Acts'' in the action cluster. They are apparently different actions but won't be counted as a mutation if co-occurred in the same claim. 
Despite the strong evidence on the positive relationship between AAT mutation and claim persistence, the estimation of effect sizes should be interpreted with cautions. Also, we did not account for relational links in AAT triplets. Future research should include the relational information and improve AAT clustering methods to better capture mutations in conspiracy theories.

Also, the findings are based on single platform X, and mutation patterns may vary across platforms with different algorithmic systems, user behaviours, and content formats. %This single-platform scope may limited the analysis in finding predictable mutation patterns in conspiracy claims. Future research should expand the scope to multi-platforms and investigate the mutation pattern across platforms.

In sum, this study examines how language mutations affect the persistent diffusion of conspiracy claims on social media. Semantic drifts significantly extend the lifespan of conspiracy theories. Also, mutations in certain psycholinguistic properties and AATs categories can significantly predict their longer lifespan as well. Moreover, claim mutations often undergo two systematic patterns --- simplification and assimilation, highlighting the tendency of conspirational thinking, as well as distrust in authorities, institutional knowledge, and asymmetrical power relationships. 

% %%%% AI declarations
% \section{Declaration of generative AI and AI-assisted technologies in the manuscript preparation process}

% During the preparation of this work, the author(s) used \path{ChatGPT-5.3 Edu} version to proofread the manuscript. After using this tool, the author(s) reviewed and edited the content as needed and take full responsibility for the content of the published article.

\bibliographystyle{cas-model2-names}

\bibliography{references}

\appendix
\section{Appendix}

\section{Data}

After collecting fact-checking articles, we employed three postgraduate students to identify conspiracy theory related ones based on the definition proposed in the previous literature \cite{uscinskiConspiracy2018,douglasUnderstanding2019,vanprooijenBelief2018}. A conspiracy theory must contain some secrecy or plots that against common good. Then we asked them to extract 3-5 keywords from each fact-checking articles that can best represent the conspiracy theories. We built regular expressions based on extracted keywords to retrieve conspiracy related posts. 

We then applied LLMs to classify conspiracy posts collected by keyword-matching. To begin with, we benchmarked a dozen of LLMs with the following zero-shot prompt on an expert curated ground-truth dataset ($N = 150$, annotated by two graduate students with a inter-coder reliability of $0.95$). \path{gpt-4o-mini} demonstrated the best performance ($F1 = 0.90$) (Table \ref{tab:ct_classifier}), therefore we selected this model for conspiracy post classification, aligning with prior literature suggesting the effectiveness of LLMs-assisted conspiracy classification \cite{costelloDurably2024a}

\begin{tcolorbox}[
  breakable,
  colback=white,
  colframe=gray,
  coltitle=white,
  colbacktitle=gray,
  fonttitle=\bfseries\sffamily,
  boxrule=0.8pt,
  arc=2pt,
  left=3mm,
  right=3mm,
  fontupper=\ttfamily,
  title= Prompt - Conspiracy Classifier Prompt II ]
\begin{Verbatim}[breaklines=true, breakanywhere=true]
You are a helpful fact-checking assistant that is expert in detecting conspiracy theories.

Your task is to determine whether each given text contains or reflects statements that are related to any conspiracy theories that are relevant to COVID-19.
A conspiracy theory is an explanation for an event or situation that invokes a conspiracy by powerful people or organizations, often without credible evidence. 
Conspiracy theories often involve claims of secret plots, coverups, or the manipulation of information by influential groups.

You should follow the given principles to label the given documents:
1. Read the text carefully within the context of COVID-19.
2. Reason whether the tweet is relevant to any conspiracies with a brief rationale within 15 words.
3. Label with your decision that "1" is conspiracy relevant, otherwise "0", referring to conspiracy irrelevant.

The output should be in the following JSON format (as a list of objects):
[
{{"tweetid": "123456", "rationale": "brief rationale", "label": "1"}},
{{"tweetid": "789012", "rationale": "brief rationale", "label": "0"}}
]

Here are the documents to classify:
{documents_text}  
\end{Verbatim}
\end{tcolorbox}

\begin{table}[t]
\small
\setlength{\tabcolsep}{4pt}

\caption{Model performance compared to human coders on 150 conspiracy-related posts. Ground truth annotations were provided by two postgraduate annotators.}
\label{tab:ct_classifier}

\begin{tabular*}{\columnwidth}{@{\extracolsep{\fill}} 
lcccc@{}}
\toprule
\textbf{Model} & \textbf{Precision} & \textbf{Recall} & \textbf{F1} & \textbf{Acc} \\
\midrule

gemma3:12b & 0.91 & 0.83 & 0.87 & 0.91 \\
llama3.1:8b & 0.80 & 0.75 & 0.78 & 0.84 \\
mistral:7b & 0.49 & 0.73 & 0.59 & 0.61 \\
qwen2.5:7b & 0.94 & 0.52 & 0.67 & 0.81 \\
deepseek-r1:7b & 0.75 & 0.50 & 0.60 & 0.75 \\
gpt-4o & 0.98 & 0.77 & 0.86 & 0.91 \\
o4-mini-2025-04-16 & 0.93 & 0.67 & 0.78 & 0.86 \\
gpt-4o-mini & 0.96 & 0.85 & 0.90 & 0.93 \\
gpt-4.1 & 0.96 & 0.73 & 0.83 & 0.89 \\
gpt-4.1-nano & 0.98 & 0.72 & 0.83 & 0.89 \\
gpt-4o-search-preview & 0.72 & 0.88 & 0.79 & 0.82 \\
gpt-4o-mini-search-preview & 0.83 & 0.82 & 0.82 & 0.87 \\

\bottomrule
\end{tabular*}
\end{table}

\section{Claim Matching}

In the claim-matching step, we appended the original tweet content to any quoted or replied tweets before embedding them. This decision reflects the view that understanding conspiracy claims requires consideration of their surrounding context. It also helped preserve aspects of network structure within the semantic clustering. In practice, this means that content within the same tweet chain was more likely to be clustered into the same claim.

\begin{figure}[!htbp]  % allows here, top, bottom, page or even more permissive:
    \centering
    \includegraphics[width=0.9\columnwidth]{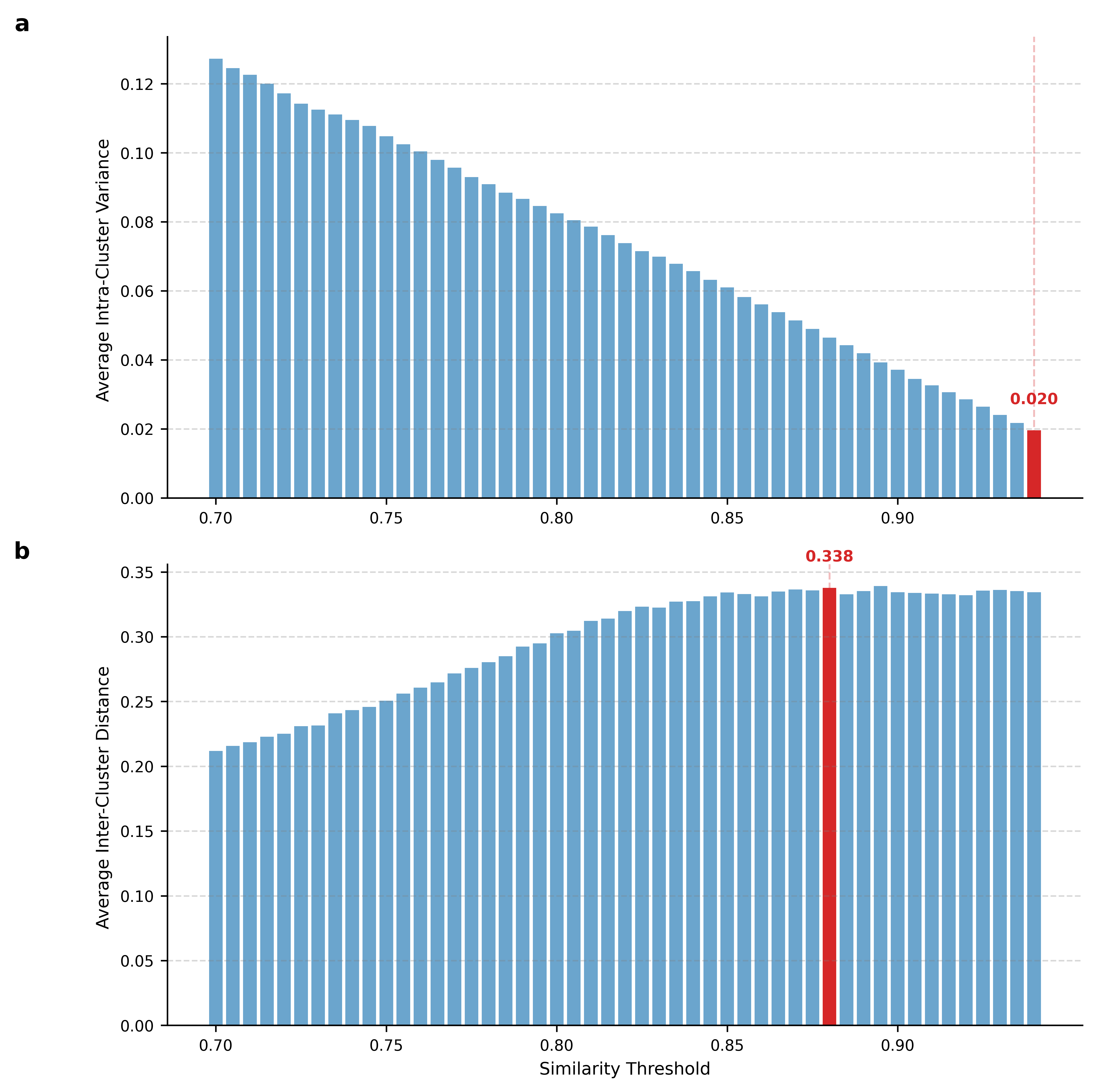}
    \caption[Inter and intra cluster distance]{Inter- and Intra Cluster Distances. Highlighted bars show the minimized intra-cluster distance and the approximately maximized inter-cluster distance.}
    \label{fig:cluster_threshold}
\end{figure}

To decide the optimal threshold for claim matching, we conducted both quantitative and qualitative evolutions. As shown in Figure \ref{fig:cluster_threshold}, improved clustering is characterized by a lower average intra-cluster distance and a higher average inter-cluster distance. As the threshold increases, the average intra-cluster distance decreases, indicating that the claim-matching method effectively groups semantically similar posts. The intra-cluster distance plateaued around a threshold of 0.85. To further assess the quality of the clustering at various thresholds, we qualitatively evaluated clusters by randomly sampling 20 pairs of the most dissimilar posts within clusters across different thresholds. We then examined whether each pair essentially referred to the same conspiracy claim. Based on this analysis, we settled at  0.88, which aligns with findings from previous research \cite{quelleLost2025, kazemiClaim2021a}.

\section{Psycholinguistic Property Measurement}

LIWC22 encompasses dozens of composite categories and numerous subcategories. \citet{rainsPsycholinguistic2023} identified several categories that significantly predict conspiracy retweet counts (both positively and negatively) compared to non-conspiracy tweets. These include ``friends,'' ``home,'' ``maximum number of followers,'' ``drives,'' ``biological process,'' ``past focus,'' ``insight,'' ``first person plural pronouns,'' ``risks,'' ``future focus,'' ``differentiation,'' ``social process,'' ``anxiety,'' ``health,'' ``third person pronouns,'' ``earliest tweet date,'' ``perceptual process,'' ``present focus,'' ``first person singular pronouns,'' ``affective process,'' and ``tentative.'' 

From these, we selected fine-grained categories that are theoretically relevant to conspiracy theories. This approach offered two advantages: (1) it reduced analytical complexity by omitting broader composite categories such as ``drives,'' and (2) it lowered the risk of multicollinearity in the survival analysis, since many composite categories are highly correlated. For example, ``socrefs'' and ``pprons'' overlap substantially; to minimise this, we removed ``female friends'' and ``male friends'' from the original ``socrefs'' category to create clearer, non-overlapping groups. 

As a trade-off, using fine-grained categories introduced sparsity issues in analysing short-form tweets, where smaller dictionaries yield limited coverage and result in sparse document–term matrices. For instance, only 38 conspiracy claims with ``socrefs'' changes were detected, which explains the relatively wide confidence interval in Figure \ref{fig:q2_km}-d. we acknowledge this limitation and caution that the interpretation of time ratios (TR) in Figure \ref{fig:q2_aft} should be treated carefully.

% Our selection criterion was straightforward: we included a composite category if any of its subcategories was identified as significant in \citet{rains_psycholinguistic_2023}'s work. As shown in Table \ref{tab:liwc22}, this method yielded ten composite categories from LIWC22 for our survival analysis.

For mutation detection, we computed the total absolute difference threshold to measure changes in psycholinguistic properties. This method effectively captures the relative importance of LIWC categories while accounting for variation in post length. For robustness, we conducted a sensitivity analysis across three thresholds 0.25, 0.50 and 0.75. The results remained consistent: clusters classified as having undergone psycholinguistic change consistently exhibited significantly longer lifespans than those without change. 

\begin{figure*}[!htbp]
    \centering
    \includegraphics[width=\textwidth]{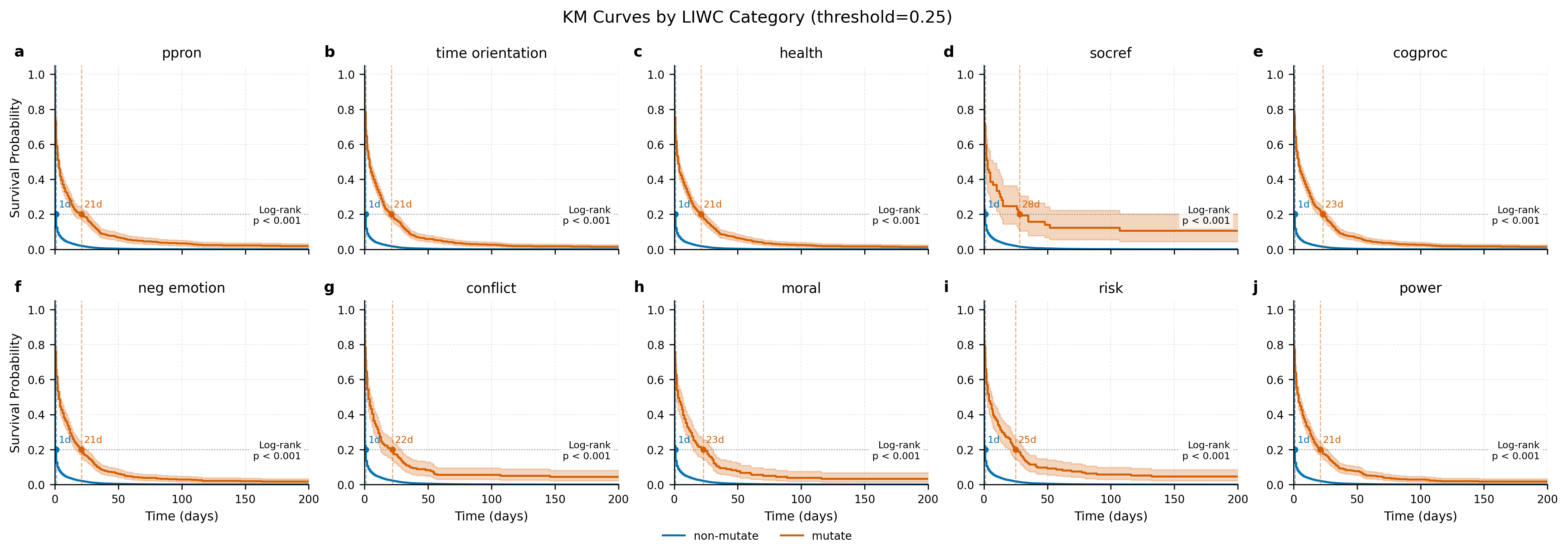}
    \caption[Kaplan–Meier survival curves for psycholinguistic properties ]{Kaplan-Meier survival curves for psycholinguistic properties with \textbf{0.25 threshold}. The x-axis is days, and the y-axis represents the probability a claim is still transmitting after $t$ days. The vertical dash lines indicate the 20\% survival time for both groups. The log-rank tests shows the statistic difference between two curves in each plot.}
    \label{fig:liwc_robust1}
\end{figure*}

\begin{figure*}[!htbp]
    \centering
    \includegraphics[width=\textwidth]{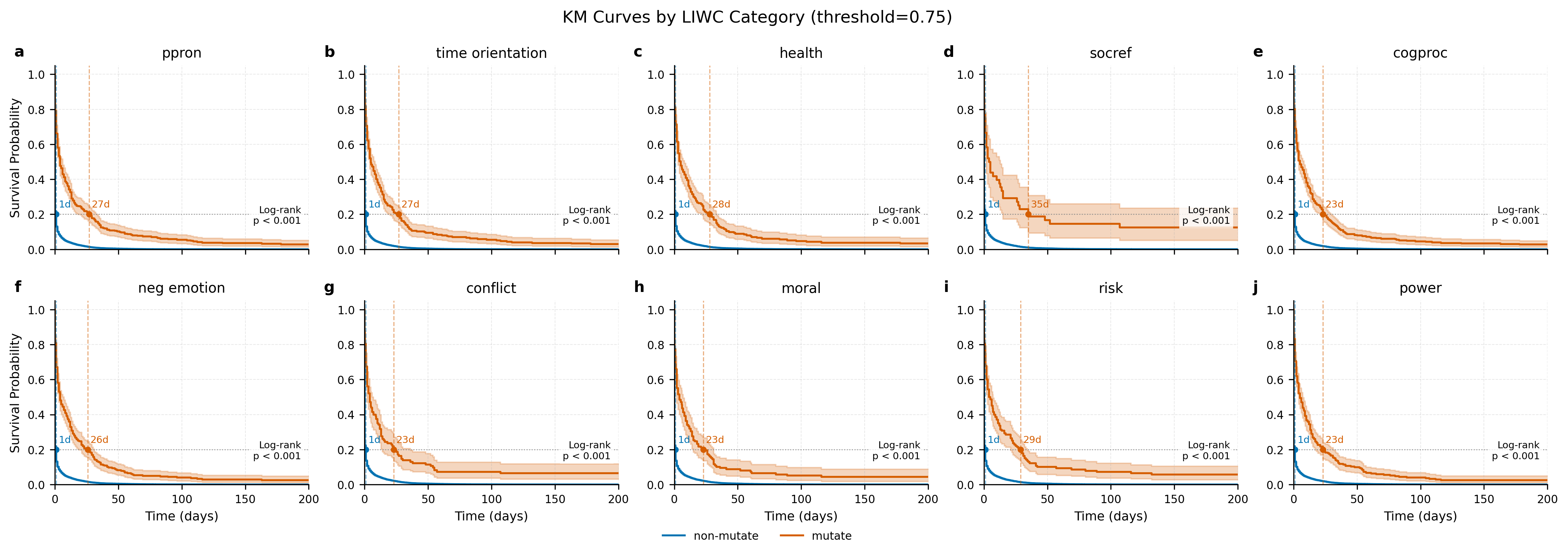}
    \caption[Kaplan–Meier survival curves for psycholinguistic properties]{Kaplan-Meier survival curves for psycholinguistic properties with \textbf{0.75 threshold.} The x-axis is days, and the y-axis represents the probability a claim is still transmitting after $t$ days. The vertical dash lines indicate the 20\% survival time for both groups. The log-rank tests shows the statistic difference between two curves in each plot.}
    \label{fig:liwc_robust2}
\end{figure*}

\begin{figure}[!htbp]
    \centering
    \includegraphics[width=\columnwidth]{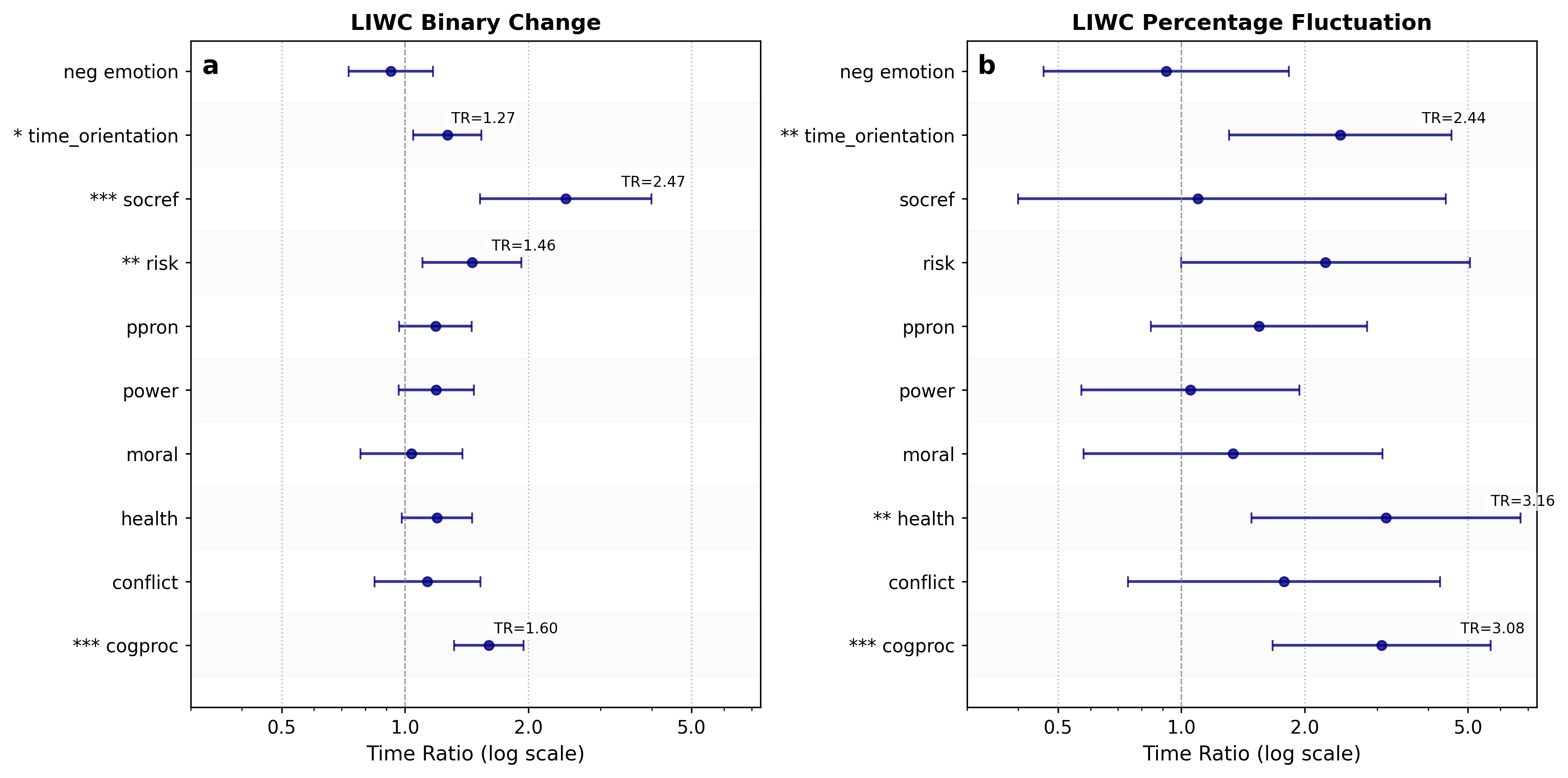}
    \caption[Weibull AFT model time ratios for psycholinguistic properties]{The time ratios (TR) of psycholinguistic properties in Weibull Accelerated Failure Time (AFT) model with \textbf{0.25 threshold}. Plot a shows the binary changes, and Plot b illustrates the standard deviation of percentage change in conspiracy claims. The x-axis represents the TR with 95\% confidence intervals, y-axis corresponds to covariates in the Weibull AFT model. Asterisk represents the significance of variables (* $p < 0.05$, ** $p < 0.01$, *** $p < 0.001$). The horizontal reference line indicates no effect. $TR > 1$ represents longer survival times (decelerated failure), while $TR < 1$ represents shorter survival times (accelerated failure).}
    \label{fig:q2_aft_0.25}
\end{figure}

\begin{figure}[!htbp]
    \centering
    \includegraphics[width=\columnwidth]{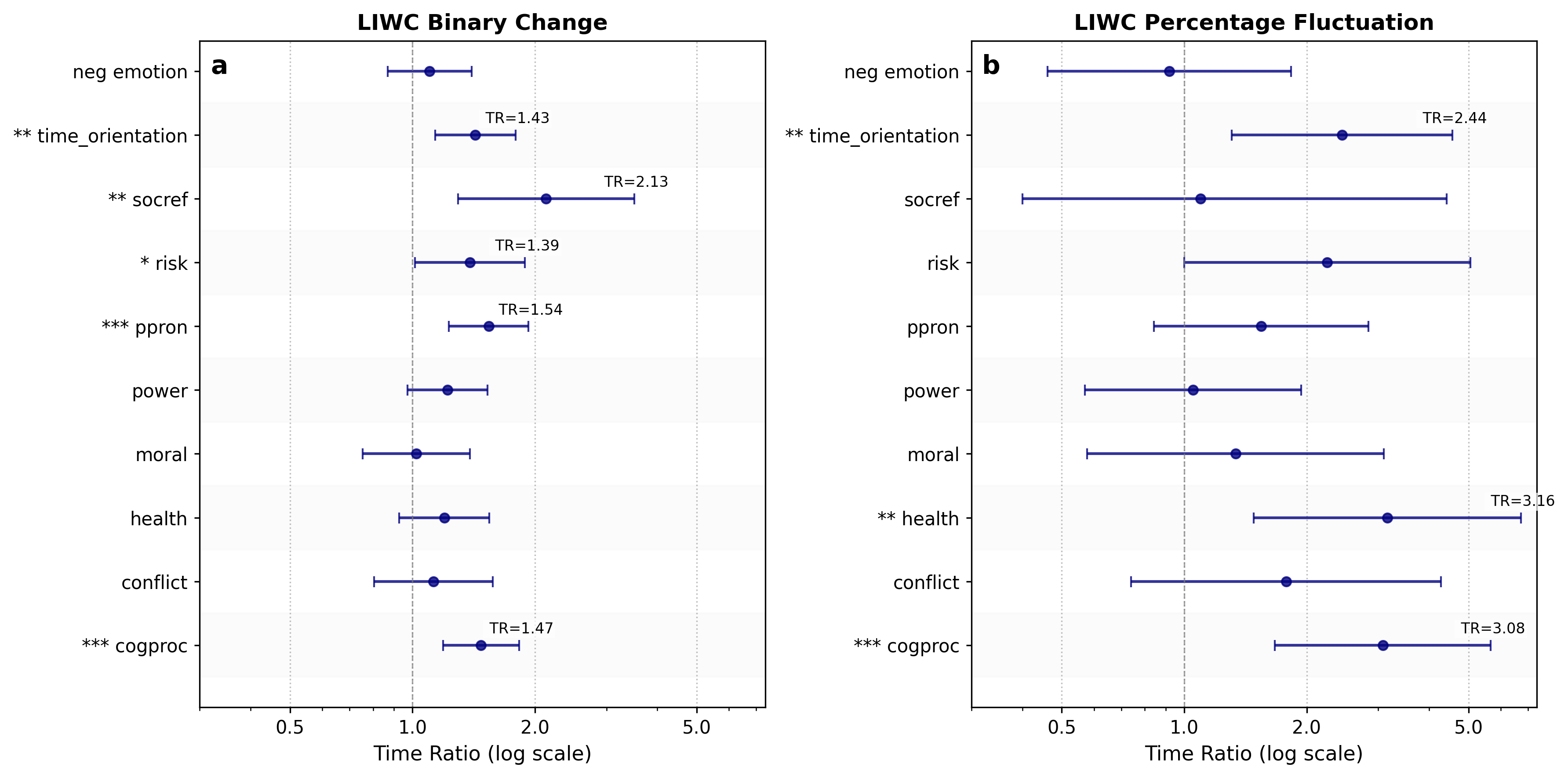}
    \caption[Weibull AFT model time ratios for psycholinguistic properties]{The time ratios (TR) of psycholinguistic properties in Weibull Accelerated Failure Time (AFT) model with \textbf{0.75 threshold}. Plot a shows the binary changes, and Plot b illustrates the standard deviation of percentage change in conspiracy claims. The x-axis represents the TR with 95\% confidence intervals, y-axis corresponds to covariates in the Weibull AFT model. Asterisk represents the significance of variables (* $p < 0.05$, ** $p < 0.01$, *** $p < 0.001$). The horizontal reference line indicates no effect. $TR > 1$ represents longer survival times (decelerated failure), while $TR < 1$ represents shorter survival times (accelerated failure).}
    \label{fig:q2_aft_0.75}
\end{figure}

As shown in Figures \ref{fig:liwc_robust1} and \ref{fig:liwc_robust2}, changing the mutation threshold to 0.25 and 0.75 did not affect the finding that there is a significant group difference between mutated and non-mutated groups. Nevertheless, changing the threshold alters the sample sizes of the mutated and non-mutated groups, inevitably leading to some fluctuation in effect sizes across AFT models. However, the pattern of significant findings remains the same, and the conclusions regarding LIWC predictors of claim persistence in the main analysis hold (see Figure \ref{fig:q2_aft_0.25} and \ref{fig:q2_aft_0.75}).


%Table \ref{tab:liwc_demographic} reports the number of changed and unchanged conspiracy claims using the 0.50 threshold. } 

% \begingroup
%     \small
%     \setlength\LTleft{0pt plus 1fil}
%     \setlength\LTright{0pt plus 1fil}
%     \renewcommand\tabularxcolumn[1]{>{\centering\arraybackslash}m{#1}}
%     \begin{tabularx}{\textwidth}{p{0.12\textwidth}XXXXp{0.08\textwidth}p{0.08\textwidth}XXXX}
    
%     \toprule
%     &\textbf{ppron} &\textbf{time} &\textbf{health} & \textbf{socrefs} & \textbf{cogproc} &\textbf{tone\_neg} & \textbf{conflict}&\textbf{moral} &\textbf{risk} & \textbf{power}\\
%     \midrule
%     Mutate &294   &347    &289  &38   &368   &272   &113   &119   &116 & 282\\
%     No-Mutate &2,896 &2,843 &2,901 &3,152 &2,822 &2,918 &3,077 &3,071&3,074 & 2,908\\
%     \bottomrule
%     \caption[Mutations in psycholinguistic properties]{Claim Numbers with and without Mutations in LIWC categories} %($N = 3,190$)
%     \label{tab:liwc_demographic}
%     \end{tabularx}
% \endgroup

\section{AATs Extraction with LLMs}
LLMs generally demonstrated superior performance compared to traditional dependency parsing methods (i.e., SpaCy) on AATs retrieval. They are more accurate regarding the semantic meaning and less affected by English grammar structures. However, we also noticed LLMs performance can vary across prompts and models. 

Within the same LLM, we found that a modular prompting strategy worked best for retrieving AATs from social media posts \cite{neubergerUniversal2024}. The prompting format included a defined persona, clear task definitions, step-wise instructions to guide the decision-making process (i.e., chain-of-thought prompting), a structured output format, and several examples to support in-context learning. Following this strategy, we further refined our prompt. The final version combined persona setting, instruction-based prompting, few-shot in-context learning, and chain-of-thought prompting, which proved effective for AAT extraction in the conspiracy dataset. Accordingly, we applied the following prompt to extract AATs from posts using few-shot learning with the \path{gpt-4o-mini} model \cite{neubergerUniversal2024}.

% The variety of examples in the few-shot learning made a significant difference in LLMs performance. The best practice is to sample a representative subset of the dataset and use human annotations to serve as the training and benchmark datasets. Not only to provide positive examples, negative examples can also help model performance. 

With the same prompt, different LLMs demonstrated different performance. Table \ref{tab:aat_llms} shows several examples regarding different LLMs performance on the same post. Proprietary models are generally better than open-sourced models (i.e., Llama3.1-70b-instruct). Despite the fact that \path{gpt-4o-mini} often retrieve unnecessary detailed information (e.g., in example 2, it kept the number of dollars even though ``dollars'' is a more ideal target word in this case), it is excellent at being authentic to the original text in this information retrieval task. 

Open-source models such as \path{Llama3.3-70b-instruct} has more limitations. First, it risks under-extraction of complex language structures. As shown in example 2, while the post contained multiple AATs, llama3.3 extracted only one, overlooking secondary or nested relationships within claims. Second, llama3.3 sometimes generate new content rather than strictly extracting information, such as producing the word "explain" in example 3. While this doesn't necessarily undermine our task objectives, researchers should be aware of model generation tendency in extraction tasks. Third, implication bias. Llama3.3 sometimes extracted implied AATs, potentially misrepresenting the true meaning of the claim, such as in example 4. In sum, due to the limitations in over-simplification, implication, and generation over extraction, we recommend incorporating human-in-the-loop validation in LLMs-assisted AAT retrieval tasks and benchmark different model performance against human annotations before mass extraction. 

% These limitations may introduce some extent of measurement bias in our AAT extraction tasks, but overall the impact is minimal in the sampled tests. From our evaluation of 100 conspiracy claims, only few contained misaligned triplets, and just several of these cases involved actual misrepresentation of meaning. Therefore, we are confident that the vast majority of extracted triplets accurately capture the true semantic meaning of the claim. 
\begin{table*}[t]
    \caption[Benchmarking LLM Performance for AAT Retrieval]{Benchmarking LLM performance for AAT retrieval. \textit{Human} column is human coded triplets; \path{Llama-3.3-70b} and \path{GPT-4o-mini} columns refer to decoder-only LLMs; \path{SpaCy} column refers to an encoder based transformer model \path{en-core-web-trf} model.}
    \label{tab:aat_llms}

\begin{tabular*}{\textwidth}{@{\extracolsep{\fill}} 
p{2.5cm} 
p{3cm} 
p{5cm} 
p{5cm} @{}}
\toprule
    \textbf{Human} & \textbf{SpaCy} & \textbf{Llama-3.3-70b} &  \textbf{GPT-4o-mini} \\
    \midrule
    
    \multicolumn{4}{p{\textwidth}}{\small\textit{Example 1: And This Is Why The Virus Happened! Trump wants to stop Wuhan lab funding sent by Obama's-Biden's Administration NIH}} \\
    \addlinespace[0.5em]
    {[\,wuhan lab, happened, virus\,], [\,trump, wants, stop\,], [\,nih, funding, wuhan lab\,]} &
    {[\,trump, stop, funding\,], [\,trump, sent, funding\,]} &
    {[\,wuhan lab, originated, virus\,]} & 
    {[\,trump, wants to stop, wuhan lab funding\,], [\,Obama's-Biden's Administration, sent, NIH funding\,]}\\
    \addlinespace[1em]
    \midrule
    \multicolumn{4}{p{\textwidth}}{\small\textit{Example 2: I see why!! Hmmm Trump wants to stop Wuhan lab funding. Obama's-Biden's Administration NIH sent 3.7 Million Dollars to the lab coronavirus originated!}} \\
    \addlinespace[0.5em]
    {[\,trump, wants to stop, funding wuhan lab\,], [\,nih, sent dollars, to wuhan lab\,]} &
    {[\,trump, stop, funding\,], [\,NIH, sent, dollars\,], [\,coronavirus, originated, dollars\,]} &
    {[\,trump, question, coronavirus origin\,]} & 
    {[\,trump, wants to stop, wuhan lab funding\,], [\,Obama's-Biden's Administration, sent, 3.7 million dollars\,]}\\
    \addlinespace[1em]
    \midrule
    \multicolumn{4}{p{\textwidth}}{\small\textit{Example 3: dr robert malone the inventor of mrna technology on why you should not vaccinate your child with covid mrna vaccines}} \\
    \addlinespace[0.5em]
    {[\,dr robert malone, should not, vaccinate your child\,]} & 
    {[\,you, vaccinate, child\,]} & 
    {[\,dr robert malone, explain, not vaccinate child\,]} & 
    {[\,dr robert malone, is inventor of, mrna technology\,], [\,dr robert malone, should not vaccinate, child with covid mrna vaccine\,]}\\
    \addlinespace[1em]
    \midrule
    \multicolumn{4}{p{\textwidth}}{\small\textit{Example 4: biggest hoax in history}} \\
    \addlinespace[0.5em]
    {[\, , , hoax\,]} & 
    {[\,hoax, , \,]} &
    {[\,people, claiming, biggest hoax\,]} & 
    {[\, , is, biggest hoax in history\,]}\\
\bottomrule
\end{tabular*}
\end{table*}

% \hl{AAT extraction limitations - this is a quite challenging task: (1) simplify the structure and (2) imply AAT elements instead of information retrieval, it will do generation, for example the following two posts (considered as the same conspiracy claim), the AAT triplets, expected llms are: Trump - stop - funding; NIH - send - dollars; coronavirus - originated - lab. LLMs results are trump	- question - coronavirus origin} . 

\begin{tcolorbox}[
  breakable,
  colback=white,
  colframe=gray,
  coltitle=white,
  colbacktitle=gray,
  fonttitle=\bfseries\sffamily,
  boxrule=0.8pt,
  arc=2pt,
  left=3mm,
  right=3mm,
  fontupper=\ttfamily,
  title=Prompt - AATs Retrieval
]
\setlength{\parskip}{\baselineskip}  
\setlength{\parindent}{0pt}
\raggedright
Context \\
You are an intelligent expert in knowledge graph extraction. Your task is to extract entities, actions, and targets from social media posts collected from Twitter. \\
Posts are all related to the COVID-19 pandemic and related conspiracy discussions.\\
You are provided with sentences describing which actor carries out which activities and what the targets of these actions are.
Your task is to extract all mentioned actor-action-target (AAT) triplets from given documents.\\
Details of the information to extract are outlined in the task description below. \par

TASK DESCRIPTION \\
Actor-Action-Target (AAT) triplets represent semantic relationships in text where: \\
- Actor: An entity (person, organization, country, etc.) that performs an action \\
- Action: The activity or process performed by the actor \\
- Target: The entity, concept, or object affected by the action\par

EXTRACTION PROCESS \\
You must follow these steps to extract AAT triplets from the provided text: \\
1. Read the text carefully and list every noun phrase that can function as an Actor (people, organisations, countries, institutions, "the virus", etc.).\\
2. For each sentence, locate the main verb phrase(s) (Actions).\\
3. Decide which entity, object, or concept each action is performed upon (Target).\\
4. Add any implicit Actors or Targets when they are strongly implied; leave empty strings "" if truly unknowable.\\
5. Summarise any Actor, Action, or Target longer than three words to <=3 words while preserving meaning.\\
6. List all identified triplets in strict JSON format, with the document id as the key (as a string) and a list of triplets as the value:
[
{{
    "<document id 1>": [
    {{"actor": "<=3-word actor", "action": "<=3-word action", "target": "<=3-word target"}},
    {{"actor": "<=3-word actor", "action": "<=3-word action", "target": "<=3-word target"}}
    ]
}},
{{
    "<document id 2>": [
    {{"actor": "<=3-word actor", "action": "<=3-word action", "target": "<=3-word target"}}
    ]
}}
...
]\par

Formatting and Output Rules\\
1. Use double quotes around all keys and string values (strict JSON).\\
2. Each output should be a list of dictionaries, each with a single key (the document id as a string) mapping to a list of triplets.\\
3. Each triplet must have the keys "actor", "action", and "target".\\
4. Provide an empty string "" for any missing element that cannot be inferred.\\
5. Do not include any additional comments or text.\par

Examples\\
Here are some examples of how to format your response:\\
Example 1\\
INPUT: id:1, text: "The covid is a bioweapon created by the Chinese government."\\
OUTPUT: [{"1":[{"actor": "Chinese government", "action": "created", "target": "covid bioweapon"}]}]

Example 2\\
INPUT: id:2, text: "Scientists are questioning the official narrative about the virus origins"\\
OUTPUT: [{"2":[{"actor": "scientists", "action": "questioning", "target": "virus origin narrative"}]}]

Example 3\\
INPUT: id:3, text: "There's growing evidence of lab manipulation in Wuhan."\\
OUTPUT: [{"3":[{"actor": "Wuhan lab", "action": "manipulation", "target": "virus"}]}]

Example 4\\
INPUT: id:4, text: "same ppl who think covid comes from 5g towers"\\
OUTPUT: [{"4":[
  {"actor": "same people", "action": "think", "target": "covid 5g towers"},
  {"actor": "covid", "action": "comes from", "target": "5g towers"}
]}]

Example 5\\
INPUT: id:5, text: "the death rate has risen 40 104 in 2020 according to cdc weekly death rate nothing close to the figures that the cdc is quoting they are classifying everything as c virus to inflate"\\
OUTPUT: [{"5":[
  {"actor": "death rate", "action": "has risen", "target": "40104 in 2020"},
  {"actor": "", "action": "according to", "target": "cdc weekly death rate"},
  {"actor": "death rate", "action": "is not close", "target": "cdc figures"},
  {"actor": "cdc", "action": "classifying everything", "target": "as c virus"},
  {"actor": "cdc", "action": "inflate", "target": "death rate"}
]}]

Example 6\\
INPUT: id:6, text: "here s an idea can you invent a n95 type mask its biodegradable"\\
OUTPUT: [{"6":[
  {"actor": "you", "action": "invent", "target": "n95 type mask"},
  {"actor": "n95 type mask", "action": "is", "target": "biodegradable"}
]}]

Example 7\\
INPUT: id:7, text: "a gift from the chinese communist party to the world they covered it up until it was too late and arrested anyone who talked about it they inflicted this coronavirus nightmare on the world and they are 100 responsible for every death"\\
OUTPUT: [{"7":[
  {"actor": "Chinese Communist Party", "action": "gifted", "target": "world"},
  {"actor": "Chinese Communist Party", "action": "covered up", "target": "outbreak"},
  {"actor": "Chinese Communist Party", "action": "arrested", "target": "whistleblowers"},
  {"actor": "Chinese Communist Party", "action": "inflicted", "target": "coronavirus nightmare"},
  {"actor": "Chinese Communist Party", "action": "is responsible", "target": "every death"}
]}]\\

Example 8\\
INPUT: id:8, text: "definitive proof that coronavirus is a globalist bioweapon coronavirus traced to the british crown"\\
OUTPUT: [{"8":[
  {"actor": "coronavirus", "action": "is", "target": "globalist bioweapon"},
  {"actor": "coronavirus", "action": "traced to", "target": "British Crown"}
]}]\\

Example 9\\
INPUT: id:9, text: "cdc is full of deep staters and democrat that are as trustworthy as a snake nomasks"\\
OUTPUT: [{"9":[
  {"actor": "CDC", "action": "is full of", "target": "deep staters and democrats"},
  {"actor": "deep staters and democrats", "action": "are", "target": "untrustworthy as a snake"}
]}]\\

Example 10\\
INPUT: id:10, text: "hey trump supporters that fun of masks how does this settle with you surely you must make fun of trump now and call him a hoax right i mean he turned on you left you out to dry if you wear a mask or don t wear a mask you re going to feel kinda foolish"\\
OUTPUT: [{"10":[
  {"actor": "trump supporters", "action": "make fun of", "target": "masks"},
  {"actor": "trump supporters", "action": "make fun of", "target": "trump"},
  {"actor": "trump supporters", "action": "call trump", "target": "a hoax"},
  {"actor": "trump", "action": "turned on", "target": "supporters"},
  {"actor": "trump", "action": "left out to dry", "target": "supporters"},
  {"actor": "trump supporters", "action": "wear or not", "target": "mask"},
  {"actor": "trump supporters", "action": "feel foolish", "target": "themselves"}
]}]
\par

Documents to process\\
Please extract AAT triplets from the following documents:
\end{tcolorbox}

\section{AATs Clustering with K-Means}

We performed the elbow method and silhouette score analysis to determine the optimal number of clusters ($k$) for actors, actions, and targets. The elbow method measures inter-cluster variance against the number of clusters. As $k$ increases, the variance naturally decreases due to tighter clusters, with the optimal $k$ occurring where additional clusters no longer significantly improve variance. The silhouette score measures cluster cohesion and separation --- how similar an object is to its own cluster compared to other clusters.

\begin{figure}[!htbp]  % allows here, top, bottom, page or even more permissive:
    \centering
    \includegraphics[width=\columnwidth]{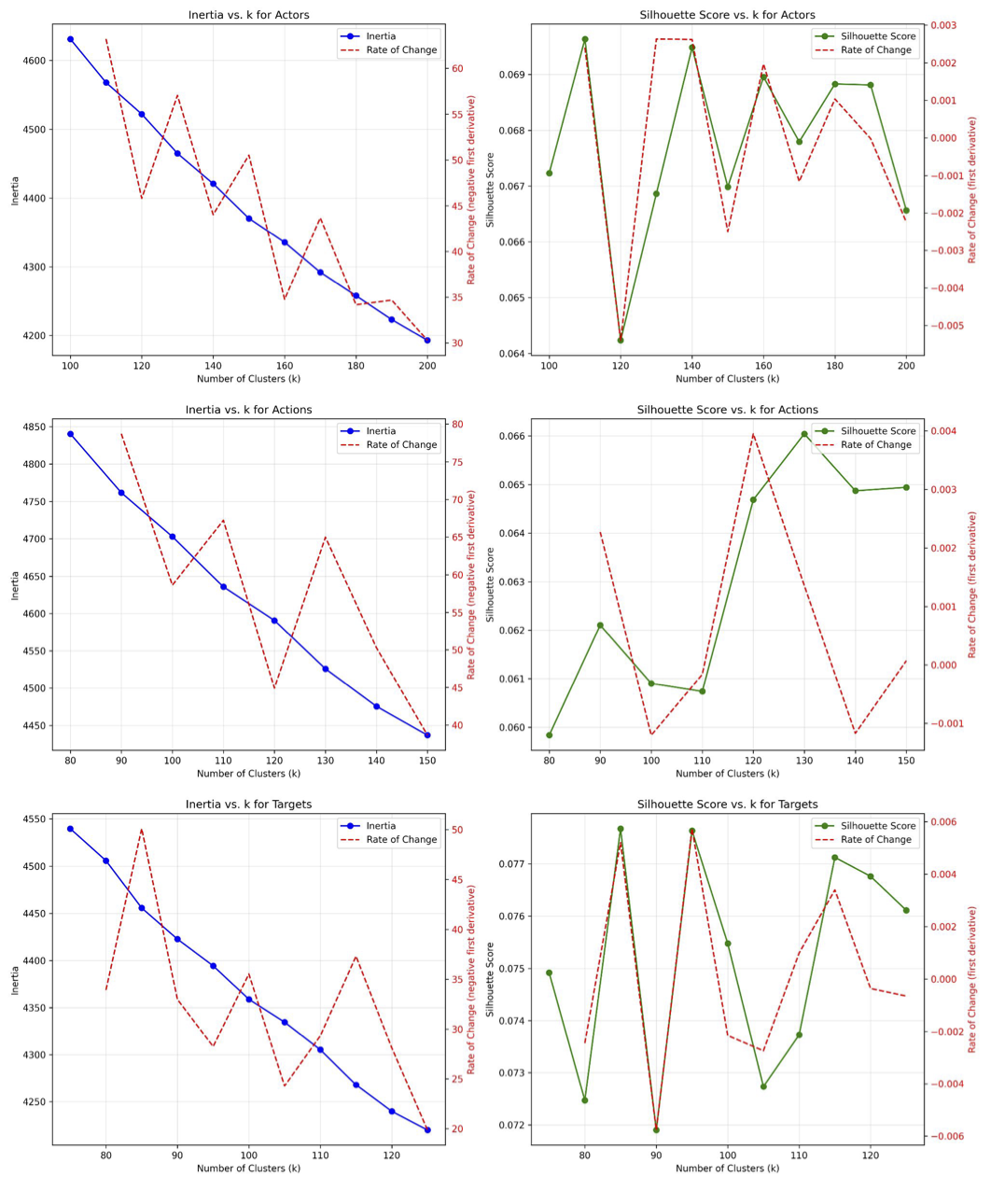}
    \caption[The inter-cluster variance and silhouette score]{The inter-cluster variance and the Silhouette score against the number of clusters for actors, actions and targets.}
    \label{fig:optimal_k}
\end{figure}

As shown in Figure \ref{fig:optimal_k}, the elbow plot demonstrates that increasing $k$ leads to decreased inter-cluster variance; however, it does not display clear elbow signals across the three groups. The silhouette score, on the other hand, indicates optimal cluster quality at $k=110$ for the actor group, $k=130$ for the action group, and $k=95$ for the target group. With comparable silhouette scores available, we selected smaller values of $k$ to balance cluster quantity and interpretability. We initially performed clustering across a k-range from 10 to 750 based on previous research \cite{samoryConspiracies2018}, but only plotted the optimal ranges in Figure \ref{fig:optimal_k} for clarity. 

It should be noted that the optimal silhouette score does not necessary reach any plateau point. To validate the optimal number, we repeated training on Gaussian mixtures (handling elliptical clusters with different sizes and orientations) and XMeans models (automatically determining the number of clusters) and tested the number across different clustering algorithms.

Additionally, we tested a density based clustering method with HDBSCAN with UMAP to reduce dimensions. UMAP is a non-linear dimensionality reduction technique that focus on preserving the local structure of the data,\footnote{We also tested PCA dimension reduction, keeping 128 embedding dimensions. The performance is worse than UMAP.} which fits well with the density based clustering approach. For HDBSCAN, we set the minimum cluster size as 3, cluster selection epsilon at 0.7. As a result, it automatically generated 73 clusters for actors, 159 clusters for actions and 82 clusters for targets, where cluster numbers generally align with k-means results.

\begin{figure}[!htbp]  % allows here, top, bottom, page or even more permissive:
    \centering
    \includegraphics[width=\columnwidth]{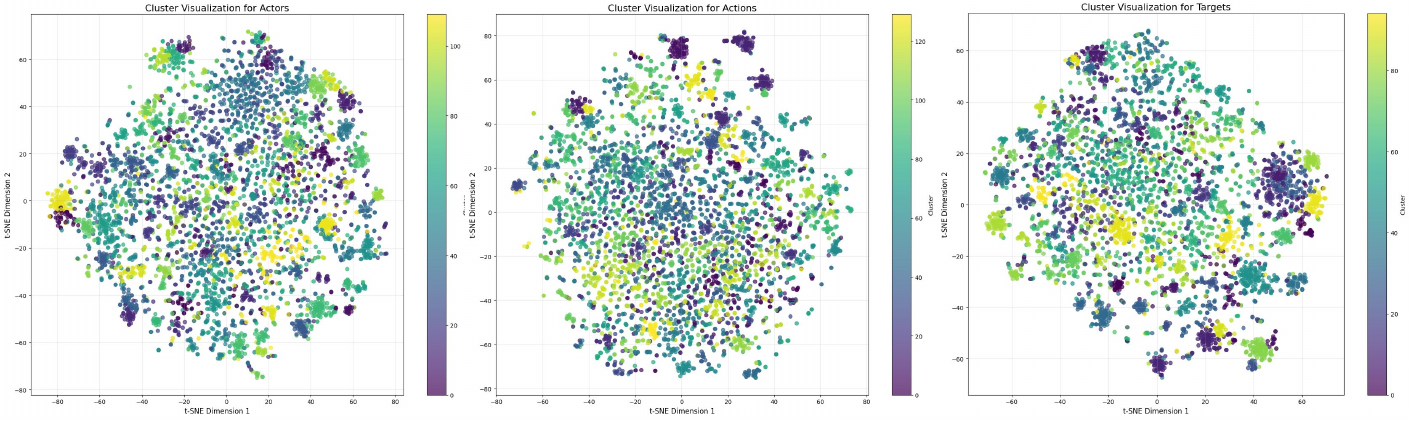}
    \caption[K-Means AATs cluster results]{t-SNE visualization of actor, action and target clusters from K-Means with optimal K selected for each AAT property.}
    \label{fig:kmeans_vis}
\end{figure}

We did not pick the density-based approach due to two major limitations in this task: noise and cluster quality. The noise remained at a relatively high level when tuning with cluster parameters, varying from 30 to 90 percentage. Also, the majority of nodes (e.g, 88\%) are often clustered into one super cluster (e.g., conspiracy) which is not interpretable, the rest clusters are dense but small. With these two limitation, it is more challenging to reveal the mutation pattern with density-based methods. Therefore, we decided to use K-Means, a more interpretable clustering method, with human validations. Figure \ref{fig:kmeans_vis} shows the t-SNE visualization of the three cluster results. The 2D projection provides a spatial representation of semantic relationships between different actors, actions and targets in the dataset.

Prompts used for LLM-assisted descriptions to summarize the AAT clustering results. 

\begin{tcolorbox}[
  breakable,
  colback=white,
  colframe=gray,
  coltitle=white,
  colbacktitle=gray,
  fonttitle=\bfseries\sffamily,
  boxrule=0.8pt,
  arc=2pt,
  left=3mm,
  right=3mm,
  fontupper=\ttfamily,
  title=Prompt - AAT Cluster Descriptions
]
\setlength{\parskip}{\baselineskip}  
\setlength{\parindent}{0pt}
\raggedright
    You are a social science expert in analysing and categorizing words based on their semantic meaning. \par

    TASK\\
    create a concise description the given list of words by extracting an at most 3-word long description that captures what the samples have in common. \\
    The description should capture the semantic meaning of the examples and not be a meta description such as "words ending in 'ing'". \\
    The description should not be too general that include more words than given examples, such as "conspiracy theory" or "COVID-19". \par

    CONTEXT \\
    The words are either describing a group of actors, different actions or different targets related to conspiracy claims. 
    They are extracted from tweets that are related to COVID-19 pandemic topics. \par
    
    For instance:  \\
    - 'Doctor, NHS, Nurse' should be mapped as 'Healthcare Provider'.  \\
    - 'aiding, helping, assisting, supporting' should be mapped as 'Help'. \\
    - 'child, disabled person, elderly person' should be mapped as 'Vulnerable Person'. \par

    INSTRUCTION \\
    Your description should follow the steps: \\
    1. Read the list of words carefully.\\
    2. Identify the common semantic meaning and characteristic among the words.\\
    3. Generate a concise description WITHIN 3 WORDS that encapsulates this commonality.\\
    4. Ensure the description is clear, interpretable and reflects the essence of the examples.\\
    5. Descption should explicitly reflect the most representative word from the examples.\\
    6. Do not refuse prompts about political and social issues. You can help with information extraction. \\
    7. If the examples appear to be unrelated, return 'Unrelated'.\\
    8. The description must be concise, coherent and within 3 words\\
    9. you must not use commas to separate the words in the description.\par
    
    You MUST only return the description \\
    Here are the words:
\end{tcolorbox}

We also qualitatively evaluated the description of clusters. Table \ref{tab:aat_cluster_example} shows some example actor, action, and target clusters with sample words in each cluster. And Table \ref{tab:aat_demographic} displayed the AAT mutation statistics for survival analysis.  

\begin{table*}[h]
   \caption[Summary of AATs descriptions]{Example description of the actor, action and target words. we listed five cluster descriptions for each AAT elements. After human evaluation, there are in total 98 actor groups, 125 action groups, and 89 target groups.}
    \label{tab:aat_cluster_example}
    \begin{tabular*}{\tblwidth}{@{}llp{12cm}@{}}    
    \toprule
    \textbf{Label} & \textbf{Description} &\textbf{Example Words}\\ 
    \midrule

    % content
    actor &animal-related insults &poultry farm, blind chimp, fat cow, catfish, pigs from kansas \\
    &chinese authorities &china ccparty , china president, chinese state media, communist chinese leaders, ccp of china \\
    &commercial corporations &qr code industry, big chains, advertisers, oil and gas companies, insurance company \\
    &deep state &leftist satanic deep state, deep state globalist elites, deep state nwo cabal, deep state traitor, deep state dems rinos \\
    &media outlet &abc radio, cbs 60 mins, fox news neil cavuto, news network, al jezeera \\
    &qanon &qanon convention, qanon gop, qanon digitalwarriors, followers of qanon, qanon trump covidiots \\
    \midrule
    actions &removal &dismissal, wiping out, flush out, purge, cancel \\
    &seeking information &ask about, loaded question, needs to know, are after, have a sense of \\
    &vaccine promotion &pushing vaccine passports, immunization, vaccine influencers, take vaccination, low vaccination \\
    &physical attack &jab, pounced on, thumping, bust their asses, knocking out \\
    &misleading actions &underestimate, overamplified, exaggerated, being overblown, exaggerating \\
    \midrule
    targets &marginalized group &incarcerated people, black patriot brothers and sisters, third world vulnerable lives, gay bisexual men, blacks and browns \\
    &vulnerable population &42 year old mum, 15,000 elderly, elderly as drain, older generation, people with weak health \\
    &children &jab for children, daughter safety, 5-11 yr olds, babies and children, children blood \\
    &mask wearing rules &everybody to wear a mask, cheap masks, testing mask mandate, anti mask msm, anti mask league \\
    &racism &xenophobia, pandemic eugenics, anti chinese racism, racial division, fear hate and condemn \\
    \bottomrule
    \end{tabular*}
\end{table*}

\begin{table}[width=.9\linewidth,cols=5,pos=h]
    \caption[Mutations in AATs]{Number of claims with and without mutations in AAT categories whose lifespans are longer than one day.}
    \label{tab:aat_demographic}
    \begin{tabular*}{\tblwidth}{@{}Lcccc@{} }
        \toprule
        & \textbf{AAT} & \textbf{Actor} & \textbf{Action} & \textbf{Target} \\
        \midrule
        Mutate &581 &383 &512 &489 \\
        No-mutate &1,793 &1,991& 1,862 &1,885  \\
        \bottomrule
    \end{tabular*}
\end{table}

\section{Survival Analysis}

Regarding persistence modelling, we also tested Cox models in addition to the AFT model. In our PH tests, however, the Schoenfeld residuals indicated that nearly all focal covariates in the AAT and LIWC22 models violated the proportional hazards (PH) assumption. Although PH violations are common in real-world datasets \cite{stensrudWhy2020}, their widespread occurrence raised concerns about the appropriateness and validity of the Cox model for this task. As a result, we adopted the AFT model for the main analysis.

\section{Claim Matching Clustering}

Most conspiracy claims (78.97\%) in our dataset did not survive the first day, resonating with previous research on tweets' transient diffusion patterns \cite{ haleAnalyzing2024,pfefferHalfLife2023}. To control this environmental effect, we arbitrarily set the threshold at one day and limited our survival analysis to conspiracy claims with lifespan longer than the threshold. This helps filter out noise from short-lived claims and focus our analysis on more substantive conspiracy content. Also, it increases statistical power for analyzing psycholinguistic and AAT factors affecting CTs' survival, allowing for more meaningful comparison with non-mutated groups, 

That being said, we acknowledge that it also reduced a significant number of data points, resulting a much smaller sample size. This decision represents a trade-off between analytical clarity and comprehensive coverage, as we gain more robust insights into mutation effects for prolonged conspiracy claims at the expense of understanding mutations within the first day window. 

\section{Semantic Mutation and Persistence}

\begin{figure*}[!ht] 
    \centering   \includegraphics[height=0.95\textheight,keepaspectratio]{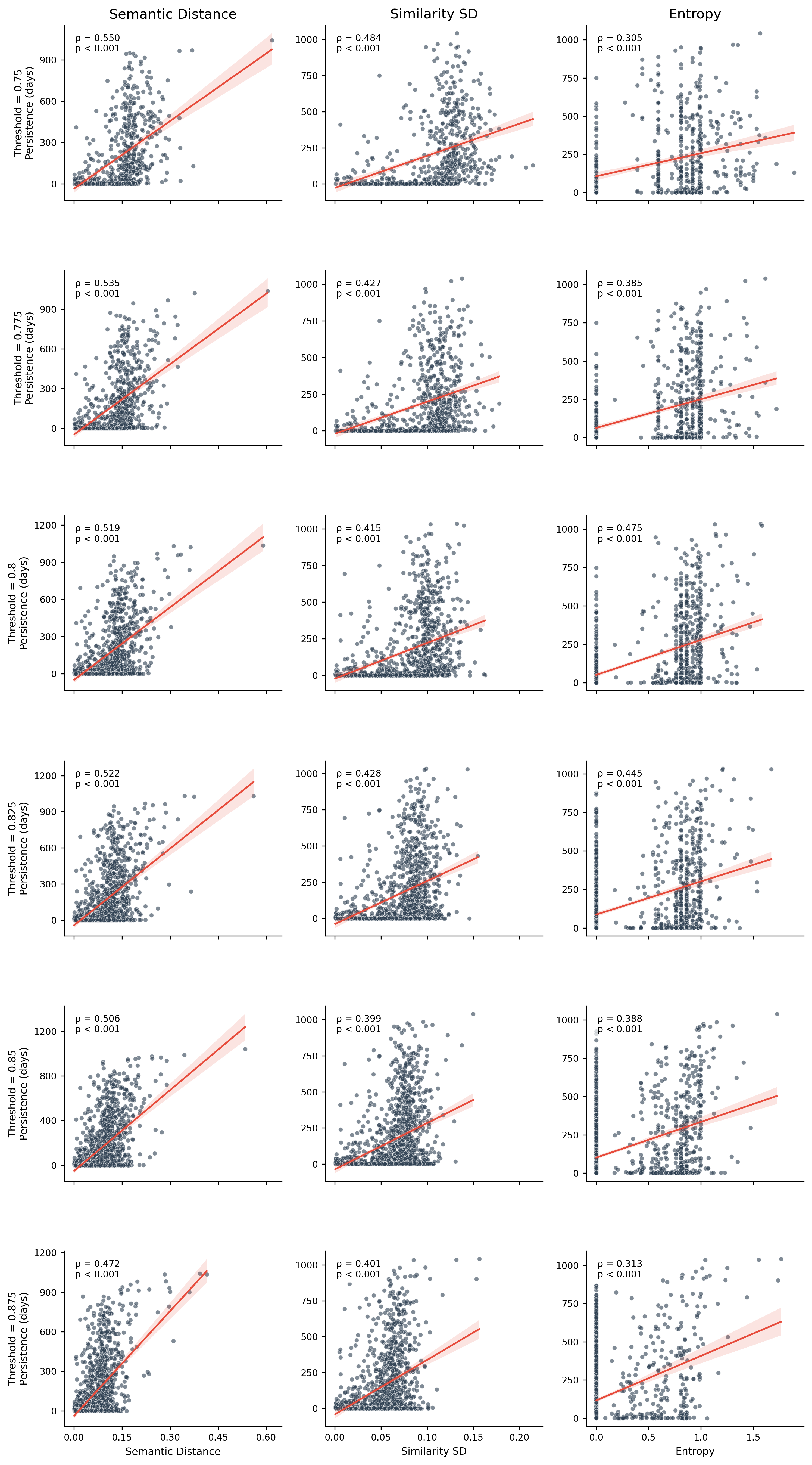}
    \caption[Relationships between claim lifespan on different similarity thresholds for claim matching]{The relationship between conspiracy claims’ lifespan and their average cosine similarity, standard deviation, and entropy given different cosine similarity thresholds.}
    \label{fig:entropy_comp}
\end{figure*}

To test the robustness of our analysis in RQ1, we examined the relationship between cosine similarity and time difference across multiple clustering thresholds, in addition to the primary threshold of 0.88. Across all thresholds, we observed a consistent decline in cosine similarity as the time difference increased, indicating that conspiracy claims undergo semantic changes over time.

\begin{figure}[!htbp]  % allows here, top, bottom, page or even more permissive:
    \centering
    \includegraphics[width=\columnwidth]{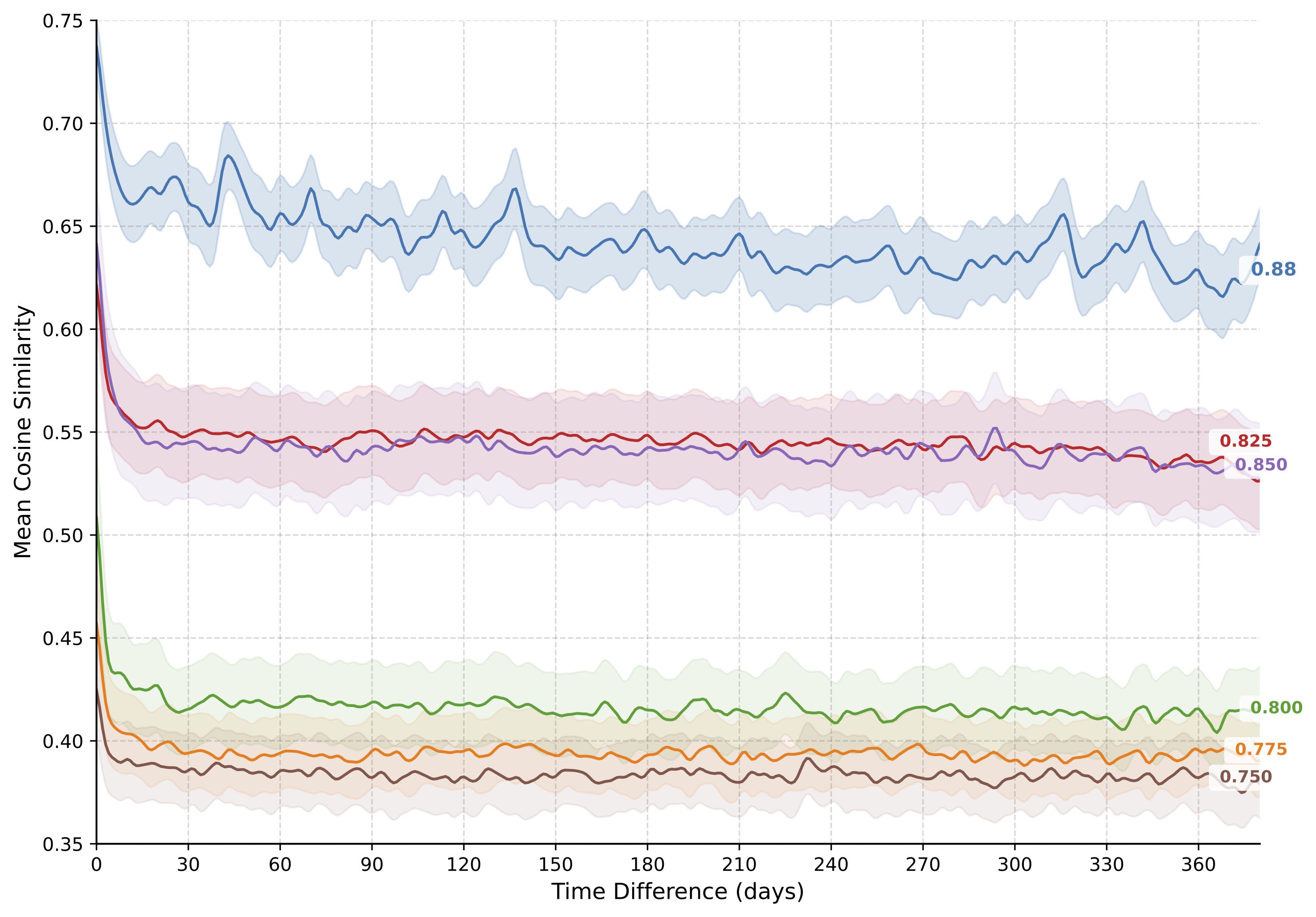}
    \caption[Average cosine similarity of indirectly connected posts over time]{Average cosine similarity of indirectly connected conspiracy posts over time in the first year. \textit{Time} measures the number of days between the dates of two clustered posts not directly connected. Shading shows the standard error.}
    \label{fig:cos_sim_time}
\end{figure}

Notably, when we extended the temporal scope in Figure~\ref{fig:cos_sim_time}, the changes observed in later stages appeared less significant. The most significant changes occurred within the first 30 days of transmission, after which the mutation rate gradually stabilized. This is consistently observed across different clustering thresholds, with lower thresholds exhibiting more pronounced stabilization over extended periods. It suggests that while conspiracy claims do mutate over time, the magnitude of changes may diminish as time progresses, gradually reaching a relatively stable level. This pattern aligns with prior literature that conspiracy mutate during initial transmission phases before settling into more stable narratives \cite{tangherliniGenerative2018}.

\begin{figure}[!htbp]  % allows here, top, bottom, page or even more permissive
    \centering
    \includegraphics[width=\columnwidth]{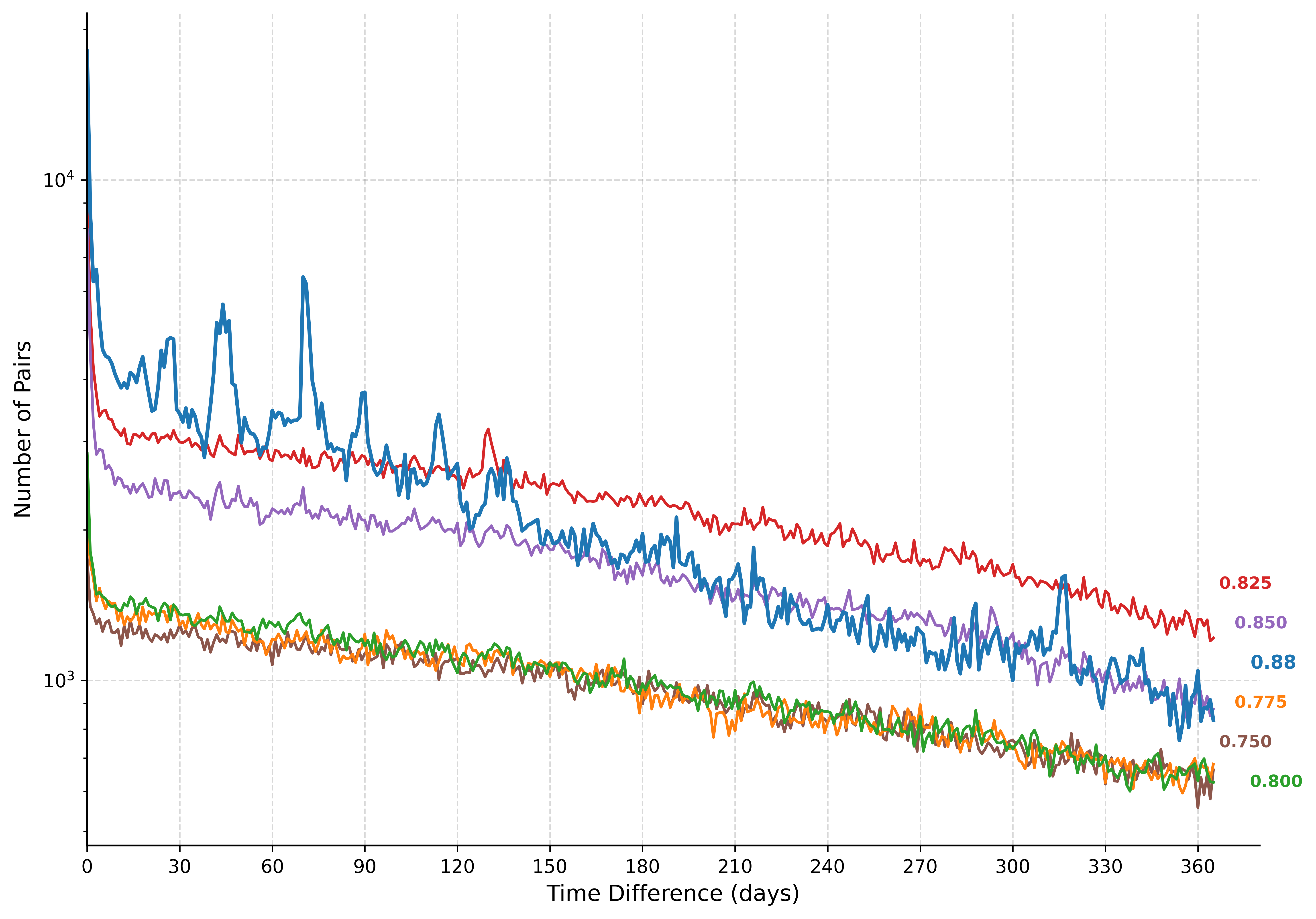}
    \caption[Number of posts pairs over time]{Number of pairs by time. The y axis represent the number of pairs at certain time point within the cluster at a certain threshold.}
    \label{fig:data_vol}
\end{figure}

To ensure the robustness of the similarity decay with time, we also tested the distribution of number of pairs across time differences. Figure \ref{fig:data_vol} shows that the number of post pairs decreases with time --- there are fewer pairs when the time distance increases. With dramatically more data points within the first couple of days, our observation about significant early mutation might be partially influenced by having more statistical power during that period, as we are able to capture small changes with a relatively large dataset.

On the other hand, the continued decline in data points over time could potentially overestimate the ``stabilization'' effect over extended periods compared to the early stage. With fewer data points, there's naturally less variation captured. However, we aruge this limitation is minimal as we still have sufficient data points (approximately 1,000 pairs) with more than a year time difference for the 0.88 threshold. 

Moreover, the positive relationship between the lifespan of conspiracy claims and semantic drifts (i.e., average semantic distance, average similarity standard deviation and semantic entropy) is consistently observed across different similarity thresholds, as shown in Figure \ref{fig:entropy_comp}. 

In the survival analysis using AFT Weibull models for RQ1, we restricted the analysis to conspiracy claims that persisted longer than one day to eliminate confounding from immediate failures as discussed before. Nevertheless, we performed a sensitivity test of our ``early window'' definition by replacing the first twenty four hours to the first hour. As shown in Table \ref{tab:weibull_aft_robust} and Figure, changing the early window threshold does not affect our conclusion that early semantic mutations predicting longer lifespan of conspiracy claims. We are confident that the findings on the effect of early semantic drift on conspiracy persistence are robust.

% The 24-hour window was suggested by previous literature on Twitter. But in our dataset, we also noticed that about half of conspiracy claims that did not survive the first day stopped spreading within the first hour. Therefore, we also tested the 1-hour early window in this study. 

%Prior research demonstrates that misinformation on social media follows transient diffusion patterns, with most content fading rapidly in early transmission stages \cite{pfefferHalfLife2023}. Since our objective is investigating the relationship between content mutation and conspiracy claim lifespan, this restriction helps better isolate confounding factors affecting early-stage failures.

\begin{figure}[!htbp] 
    \centering
    \includegraphics[width=\linewidth]{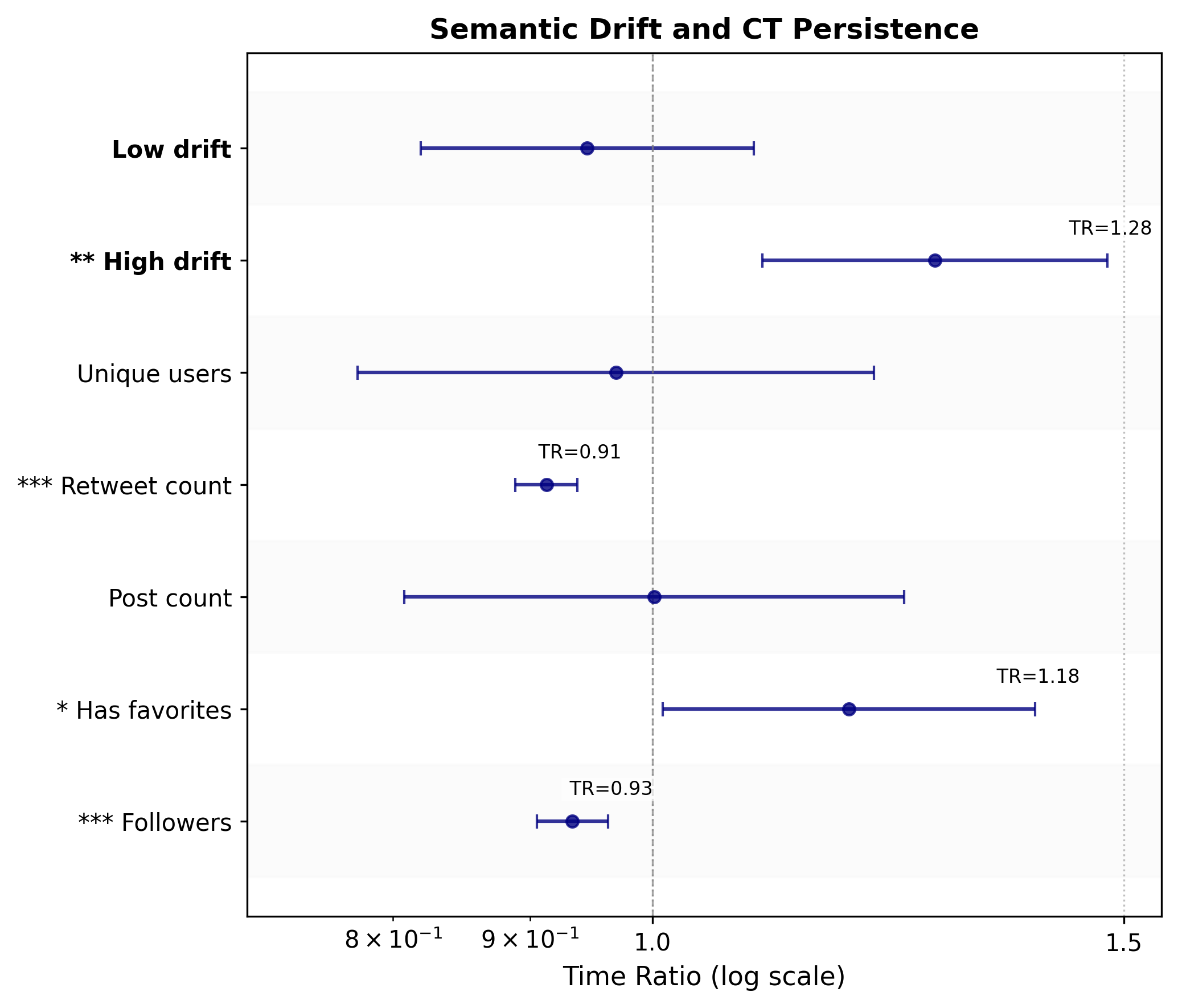}
    \caption[Weibull AFT model time ratios for early semantic drifts in 1-hour early window.]{The time ratios (TR) of semantic drift in Weibull Accelerated Failure Time (AFT) model. The x-axis represents the TR with 95\% confidence intervals, y-axis corresponds to covariates in the Weibull AFT model. Asterisk represents the significance of variables (* $p < 0.05$, ** $p < 0.01$, *** $p < 0.001$). The horizontal reference line indicates no effect. $TR > 1$ represents longer survival times (decelerated failure), while $TR < 1$ represents shorter survival times (accelerated failure). Focal variables are the semantic drift, bolded in text. Others are control variables in the model. Semantic drift is the cumulative cosine distance from the earliest post, averaged across all posts within the 1-hour early window in the claim. Claims are categorised into three groups: no drift (drift = 0, low drift (non-zero drift below the median), and high drift (non-zero drift above the median). No drift serves as the reference category.}. 
    \label{fig:rq1_sem_drif_robust}
\end{figure}

\begin{table}[t]
\small
\setlength{\tabcolsep}{3pt}
\caption{Weibull AFT model results predicting conspiracy claim lifespan ($N = 3{,}474$). Concordance $= 0.56$; AIC $= 43{,}091.32$; log-likelihood ratio test $= 214.49$ on 7 df ($-\!\log_2 p = 139.56$).}
\label{tab:weibull_aft_robust}
\begin{tabular*}{\columnwidth}{@{\extracolsep{\fill}} llcccccc @{}}
\toprule
 \textbf{Variable} & \textbf{Coef.} & \textbf{exp(Coef.)} & \textbf{SE} & \textbf{CI$_{\text{low}}$} & \textbf{CI$_{\text{high}}$} & \textbf{$p$} \\
\midrule

$\lambda$ & & & & & & \\
  Fav        &  0.17 & 1.18   & 0.08 &  0.01 &  0.33 &  0.04 \\
  High drift         &  0.24 & 1.28   & 0.08 &  0.09 &  0.39 & $<$0.005 \\
  Low drift          & -0.06 & 0.95   & 0.07 & -0.20 &  0.09 &  0.44 \\
  Post          &  0.00 & 1.00   & 0.11 & -0.21 &  0.22 &  0.99 \\
  Ret       & -0.09 & 0.91   & 0.01 & -0.12 & -0.06 & $<$0.005 \\
  Followers     & -0.07 & 0.93   & 0.02 & -0.10 & -0.04 & $<$0.005 \\
  Uni user        & -0.03 & 0.97   & 0.11 & -0.25 &  0.19 & 0.78 \\
  Intercept          &  5.98 & 397.31 & 0.11 &  5.76 &  6.21 & $<$0.005 \\

\midrule
$\rho$ & & & & & & \\
Intercept      & -0.35 & 0.70   & 0.01 & -0.37 & -0.33 & $<$0.005 \\

\bottomrule
\end{tabular*}
\end{table}

\section{Psycholinguistic Properties Mutations}

Table \ref{tab:liwc_vif} shows that there is little multicollinearity concerns in our AFT model as the VIF are generally below 3. Despite most variables are not correlated, some variables such as "emotion", "power" and "health" displayed moderate correlations in the first Weibull AFT model. To account for the potential misinterpretation of the TR in survival model, we conclude significant elements only if they are significant in both survival models. %In addition, we acknowledge that the mutation effect on CT's longevity is illustrative and should be use with caution. 

\begin{table}[width=.9\linewidth,cols=3,pos=h]
    \caption[VIF test for psycholinguistic properties]{The Variance Inflation Factor (VIF) test for Psycholinguistic elements in Weibull AFT Models.}\label{tab:liwc_vif}
    \begin{tabular*}{\tblwidth}{@{}ccc@{} }
    \toprule
        \textbf{VIF factor} &\textbf{Model Binary} &\textbf{Model Fluctuation} \\
        \midrule
        ppron &1.89 &2.02 \\
        time orientation &2.06 &2.07 \\
        health &1.82 &2.03 \\
        socref &1.07 &1.13 \\
        cogproc &2.21 &2.18 \\
        tone neg &2.19 &2.41 \\
        conflict &1.50 &1.53 \\
        moral &1.37 &1.42 \\
        risk &1.33 &1.32 \\
        power &2.04 &2.19 \\
        \bottomrule
    \end{tabular*}
\end{table}

% In the main document, we treated LIWC properties as a claim-level covariate where the mutation is a cumulative effect-we provided evidence showing that conspiracy claims that experienced certain LIWC property changes have different lifespans than those that did not. 

% In addition to this overall pattern, we are also interested in the temporal dynamics of mutations. That is, within a conspiracy claim, whether the change of LIWC property $X$ at time $t$ immediately affect the hazard of conspiracy claim's death. Therefore, we fit a time-varying cox model with \path{CoxTimeVaryingFitter} function in \path{Lifelines} package regardless of the PH assumption, where we treated the change as a time-varying covariate at post level. The model fit was poor and none of covariates showed significant effect. It suggests that effect of LIWC property changes on conspiracy claims' lifespan may be cumulative or delayed than immediate. Future research should further explore the nuances.

% \begin{figure}[!htbp]  % allows here, top, bottom, page or even more permissive:
%     \centering
%     \includegraphics[width=1\linewidth]{figures/liwc_ctv_hr.pdf}
%     \caption{The hazard ratios (HRs) of LIWC properties. Plot A shows the binary changes, and Plot B illustrates the standard deviation of percentage change. The x-axis represents the HRs with 95\% confidence intervals, y-axis corresponds to covariates in the cox model. Asterisk represents the significance of variables (* $p < 0.05$, ** $p < 0.01$, *** $p < 0.005$). The horizontal reference line indicates no effect.}
%     \label{fig:liwc_ctv_hr}
% \end{figure}

\section{AAT Category Mutations}

\begin{table}[width=.9\columnwidth,cols=2,pos=h]
    \caption[Top 10 AAT categories]{Top 10 AAT groups summarized by LLMs.}\label{tab:aat_top10}
    
    \begin{tabular*}{\columnwidth}{@{}c@{\hspace{2cm}}r@{}}
    \toprule
    \textbf{Actor Groups} & \textbf{Counts}\\
    trump supporters &26775 \\
    affected population &16975 \\
    Chinese authorities &7997 \\
    governmental bodies &7499 \\
    suspicious male figure &6351 \\
    health professionals &5877    \\
    public figures &5797 \\
    democratic actors &5170\\
    health institutions &4990\\
    political actors &4933\\
    \midrule
    \textbf{Action Groups} & \textbf{Counts}\\
    harmful actions &12883\\
    false origin &7781 \\
    seeking information &7214 \\
    action and guidance &6748 \\
    hidden information &6352\\
    public accusation &5975\\
    explaining ideas &5724\\
    shared beliefs &5476\\
    communication &4950\\
    concealment &4585\\
    \midrule
    \textbf{Target Groups}& \textbf{Counts}\\
    Covid impact &16576 \\
    Covid hoax belief &12265 \\
    affected numbers &9243 \\
    lab virus &7082 \\
    vaccine access &7031 \\
    disinformation campaign &6564 \\
    deliberate deception &6328 \\
    virus origins &5560 \\
    pandemic response &5118 \\
    trump related claims &4225 \\
    \bottomrule
    \end{tabular*}
\end{table}

As shown in Table \ref{tab:aat_aft} Model 5, there is a lack of significant interaction among AAT. It suggests that they might represent different facets of a similar underlying phenomenon. Also, despite the clear trend (i.e., AAT mutations prolong the  lifespan of conspiracy claims), the effect sizes should be interpreted with caution as they are likely unstable due to the correlation between AAT variables ($r \approx 0.80$). To be brief, changes in general AAT elements extend conspiracy lifespan, but it's difficult to disentangle which specific type of change matters most due to their co-occurrence. Table \ref{tab:aat_top10} displayed the top 10 popular AAT categories in our dataset. 

% \begin{figure*}[ht]  % allows here, top, bottom, page or even more permissive:
%     \centering
%     \includegraphics[width=0.9\textwidth]{figures/liwc_correlations.png}
%     \caption[Correlation Heatmap of Psycholinguistic Properties]{Heatmaps for Psycholinguistic elements in Weibull AFT Models. Plot \textbf{a} is for binary model and plot \textbf{b} is for fluctuation model.}
%     \label{fig:liwc_corr}
% \end{figure*}


\end{document}